\UseRawInputEncoding
\documentclass[10pt,journal,compsoc]{IEEEtran}
\usepackage{epsfig}
\usepackage{epstopdf}
\usepackage{graphicx}
\usepackage{amsmath}
\usepackage{amssymb}
\usepackage{multirow}
\usepackage{bbding}
\usepackage{ragged2e}
\usepackage{color}
\ifCLASSOPTIONcompsoc
  \usepackage[nocompress]{cite}
\else
  \usepackage{cite}
\fi
\ifCLASSINFOpdf
\else
\fi
\newcommand*{\red}{\textcolor{black}}
\begin{document}

\title{Cross-Modal Causal Relational Reasoning for Event-Level Visual Question Answering}

\author{Yang~Liu,~\IEEEmembership{Member,~IEEE}, Guanbin Li,~\IEEEmembership{Member,~IEEE},
        and~Liang~Lin,~\IEEEmembership{Senior~Member,~IEEE}
\IEEEcompsocitemizethanks{\IEEEcompsocthanksitem 

Y. Liu, G. Li and L. Lin are with the School of Computer Science and Engineering, Sun Yat-sen University, Guangzhou, China, and also with the GuangDong Province Key Laboratory of Information Security Technology. \protect
E-mail: liuy856@mail.sysu.edu.cn; liguanbin@mail.sysu.edu.cn; linliang@ieee.org. (Corresponding author: Liang Lin.)}}

\markboth{IEEE TRANSACTIONS ON PATTERN ANALYSIS AND MACHINE INTELLIGENCE, VOL. X, NO. X, XXX}%
{Shell \MakeLowercase{\textit{et al.}}: Bare Demo of IEEEtran.cls for Computer Society Journals}

\IEEEtitleabstractindextext{
\begin{abstract}
\justifying{\red{Existing visual question answering methods often suffer from cross-modal spurious correlations and oversimplified event-level reasoning processes that fail to capture event temporality, causality, and dynamics spanning over the video. In this work, to address the task of event-level visual question answering, we propose a framework for cross-modal causal relational reasoning. In particular, a set of causal intervention operations is introduced to discover the underlying causal structures across visual and linguistic modalities. Our framework, named \textbf{C}ross-\textbf{M}odal \textbf{C}ausal Relat\textbf{I}onal \textbf{R}easoning (CMCIR), involves three modules: i) Causality-aware Visual-Linguistic Reasoning (CVLR) module for collaboratively disentangling the visual and linguistic spurious correlations via front-door and back-door causal interventions; ii) Spatial-Temporal Transformer (STT) module for capturing the fine-grained interactions between visual and linguistic semantics; iii) Visual-Linguistic Feature Fusion (VLFF) module for learning the global semantic-aware visual-linguistic representations adaptively. Extensive experiments on four event-level datasets demonstrate the superiority of our CMCIR in discovering visual-linguistic causal structures and achieving robust event-level visual question answering. The datasets, code, and models are available at https://github.com/HCPLab-SYSU/CMCIR. }}
\end{abstract}

}

\maketitle

\IEEEdisplaynontitleabstractindextext

\IEEEpeerreviewmaketitle

\IEEEraisesectionheading{\section{Introduction}\label{sec:introduction}}
\IEEEPARstart{W}ith the rapid development of deep learning \cite{krizhevsky2012imagenet}, event understanding \cite{krishna2017dense}
has become a prominent research topic in video analysis \cite{zhou2018temporal,liu2021semantics,liu2022tcgl} because videos have good potential to go beyond image-level understanding (scenes, people, objects, activities, etc.) to understand event temporality, causality, and dynamics. Accurate and efficient cognition and reasoning over complex events \red{are} extremely important in video-language understanding.
Since natural language can potentially describe a richer event space \cite{buch2022revisiting} that facilitates \red{deeper} event understanding, we focus on the complex (temporal, causal) event-level visual question answering task in \red{a} cross-modal (visual, linguistic) setting. \red{Our task aims to fully comprehend the richer multi-modal event space and answer the given question in a causality-aware way}. To achieve event-level visual question answering \cite{das2017visual,anderson2018vision,zellers2019recognition}, the model needs to have a fine-grained understanding of video and language content involving various complex relations, such as spatial-temporal visual relation, linguistic semantic relation, and visual-linguistic causal dependency. Therefore, robust and reliable multi-modal relation reasoning is essential in event-level visual question answering. Actually, understanding events in multi-modal visual-linguistic context is a long-standing challenge. Existing visual question answering methods \cite{huang2020location,li2019beyond,le2020hierarchical,park2021bridge} use recurrent neural networks (RNNs) \cite{sukhbaatar2015end}, attention mechanisms \cite{vaswani2017attention} or Graph Convolutional Networks \cite{kipf2016semi} for relation reasoning between visual and linguistic modalities. Although achieving promising results, these methods suffer from two common limitations.

\begin{figure}
\begin{center}
\includegraphics[scale=0.42]{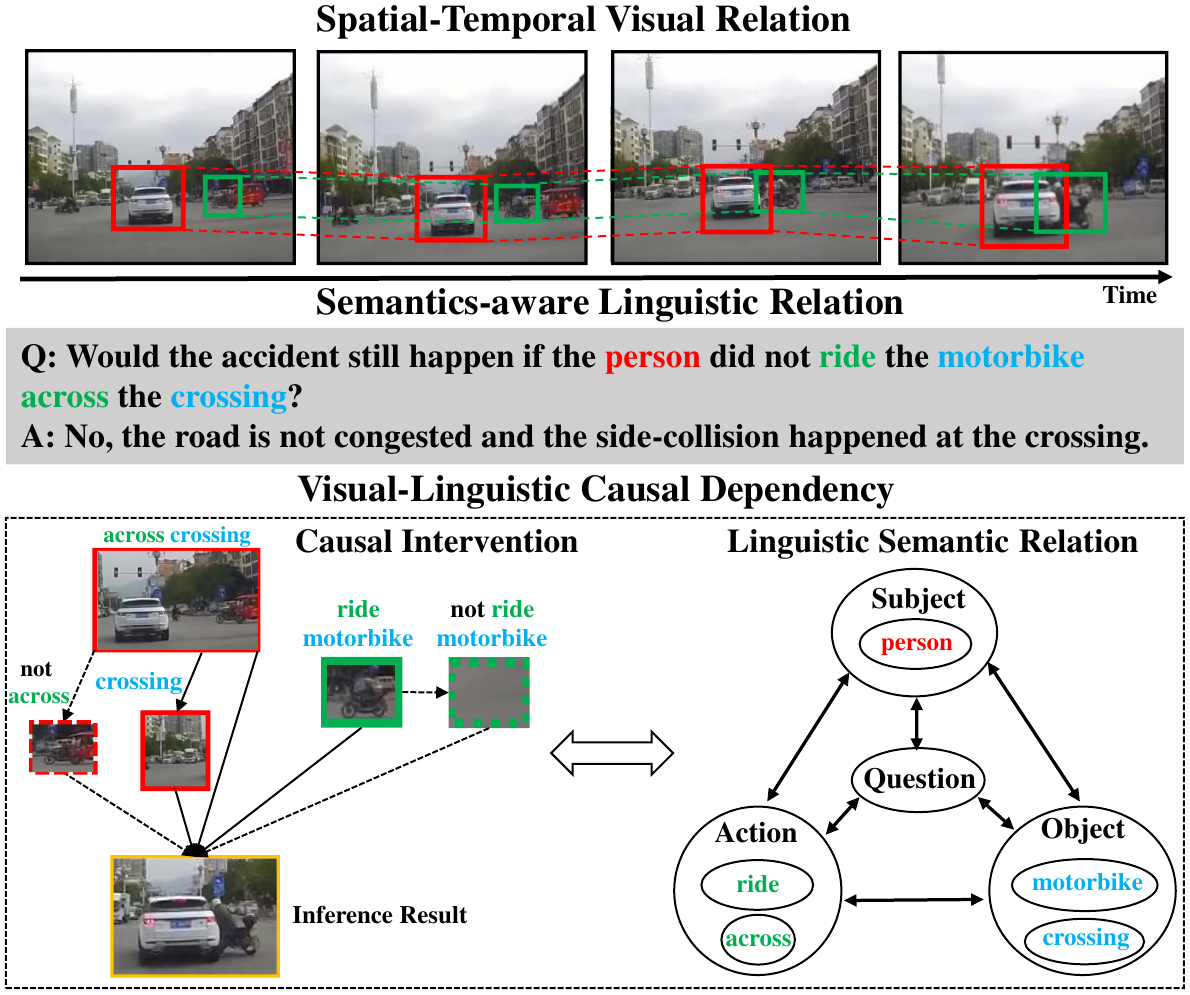}
\end{center}
    \vspace{-15pt}
   \caption{An example of \red{an} event-level counterfactual visual question answering task. The counterfactual inference is to obtain the outcome of certain hypothesis that does not occur in the visual scene. To infer the causality-aware answer, the model is required to explore the visual-linguistic causal \red{dependencies} and spatial-temporal relation.}
      \vspace{-10pt}
\label{Fig1}
\end{figure}

\begin{figure*}
\begin{center}
\includegraphics[scale=0.185]{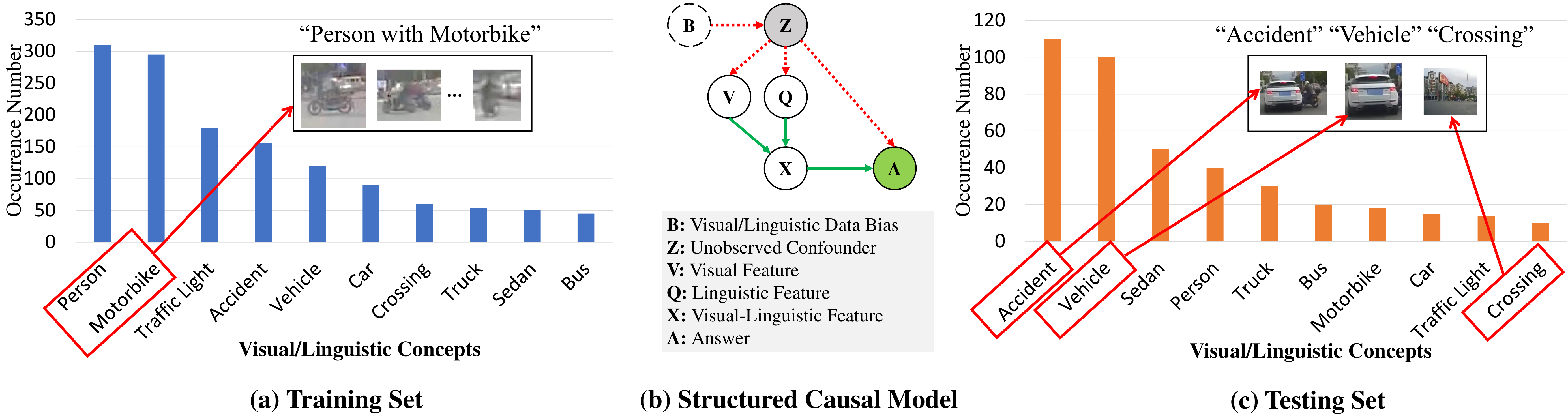}
\end{center}
    \vspace{-20pt}
   \caption{\red{The example (Fig. \ref{Fig1}) shows why the visual question answering model without causal reasoning tends to learn spurious correlations. (a) A training dataset constructed with visual and linguistic biases where the concepts ``person" and ``motorbike" frequently appear. (b) The structured causal model (SCM) shows how the confounder induces the spurious correlation in event-level visual question answering. The green path denotes the unbiased visual question answering (the true causal effect). The red path is the biased visual question answering caused by the confounders (the back-door path). (c) As a result, if we provide some samples where the ``vehicle" concept is highly related to the ``accident" to reason how the accident actually happens, the model does not really exploit the true question intention and dominant visual evidence to infer the answer.}}
   \vspace{-10pt}
\label{Fig2}
\end{figure*}

\red{First, existing visual question answering methods usually focus on simple events that do not require a deep understanding of causality, temporal relations, and linguistic interactions, and tend to overlook more challenging events. In Fig. \ref{Fig1}, given a video and an associated question, a typical human reasoning process involves first memorizing the relevant objects and their interactions in each video frame (e.g., car runs on the road, person rides a motorbike across a crossing), then deriving the corresponding answer based on this memorized video content. However, the event-level counterfactual visual question answering task in Fig. \ref{Fig1} requires the outcome of certain hypotheses (e.g., ``the person did not ride the motorbike across the crossing'') that do not occur in the given video. Simply correlating relevant visual contents cannot get the right inference result without discovering the hidden spatial-temporal and causal dependencies. To accurately reason about the imagined events under the counterfactual condition, the model must conduct hierarchical relational reasoning and fully explore the causality, logic, and spatial-temporal dynamic structures of the visual and linguistic content. This involves performing causal intervention to discover the true causal structure that facilitates answering the question truthfully based on the imagined visual evidence and the correct question intention. However, the multi-level interaction and causal relations between the language and spatial-temporal structure of complex multi-modal events are not fully explored.}

\red{Second, the current visual question answering models tend to capture spurious linguistic or visual correlations introduced by the confounders rather than the true causal structure and causality-aware multi-modal representations, leading to an unreliable reasoning process \cite{niu2021counterfactual,yang2021causal,wang2021causal,li2022invariant}. Fig. \ref{Fig2} shows that some frequently appearing concepts in linguistic and visual modalities can be considered as the confounders. The ``linguistic bias'' represents strong correlations between questions and answers, while the ``visual bias'' represents the strong correlations between certain key visual features and answers. For example, the training dataset is built with visual and linguistic biases, where the concepts ``person" and  ``motorbike" frequently appear (Fig. \ref{Fig2}). Such biased dataset entails two causal effects: visual and linguistic biases $B$ leads to confounder $Z$, which then affects the visual feature $V$, question feature $Q$, visual-linguistic feature $X$, and the answer $A$. Thus, we can draw two causal links to describe these causal effects: $Z\rightarrow\{V,Q\}\rightarrow X$ and $Z\rightarrow A$. If we want to learn the true causal effect $\{V,Q\}\rightarrow X\rightarrow A$ while using the biased dataset to train this model (Fig. \ref{Fig2} (a)), the model may simply correlate the concepts ``person'' and ``motorbike'', i.e., through $Z\rightarrow\{V,Q\}\rightarrow X$, and then use this biased knowledge to infer the answer, i.e., through $Z\rightarrow A$. In this way, this model learns the spurious correlation between $\{V,Q\}$ and $A$ through the backdoor path $A\leftarrow Z\rightarrow \{V,Q\}\rightarrow X$ induced by the confounder $Z$, as shown in Fig. \ref{Fig2} (b). Therefore, the model may learn the spurious correlation between ``motorbike"  and ``person" without considering the ``vehicle" concept (i.e., exploit the true question intention and dominant visual evidence) to reason how the accident occurred. Since the potential visual and linguistic correlations are complicated in complex events, there are significant differences in visual and linguistic biases between the training and testing sets. To mitigate the dataset bias, causal inference \cite{pearl2016causal} has shown promising performance in scene graph generation \cite{tang2020unbiased}, image classification\cite{yue2020interventional} and image question answering \cite{wang2020visual,niu2021counterfactual}. However, directly applying existing causal methods to the event-level visual question answering task may yield unsatisfactory results due to the unobservable confounder in the visual domain and the complex interaction between visual and linguistic content.}

\red{To mitigate the aforementioned limitations, this paper proposes a framework named \textbf{C}ross-\textbf{M}odal \textbf{C}ausal Relat\textbf{I}onal \textbf{R}easoning (CMCIR) for event-level VQA. The proposed Causality-aware Visual-Linguistic Reasoning (CVLR) module addresses the confounders' bias and uncovers causal structures in visual and linguistic modalities through front-door and back-door causal interventions. To address the unobservable confounder in the visual modality, a Local-Global Causal Attention Module (LGCAM) is proposed, which uses attention to aggregate local and global visual representations causality-awarely. Additionally, a back-door intervention module is designed to discover the causal effect within the linguistic modality. A Spatial-Temporal Transformer (STT) is introduced to model the multi-modal interaction between appearance-motion and language representations, containing Question-Appearance (QA), Question-Motion (QM), Appearance-Semantics (AS), and Motion-Semantics (MS) modules.  Finally, a novel Visual-Linguistic Feature Fusion (VLFF) module is proposed to adaptively fuse causality-aware visual and linguistic features. Experimental results on various datasets show that CMCIR outperforms state-of-the-art methods. The main contributions of the paper can be summarized as follows:}
\red{
\begin{itemize}
\item We propose a causality-aware event-level visual question answering framework named \textbf{C}ross-\textbf{M}odal \textbf{C}ausal Relat\textbf{I}onal \textbf{R}easoning (CMCIR), to discover true causal structures via causal intervention on the integration of visual and linguistic modalities and achieve robust event-level visual question answering performance. To the best of our knowledge, we are the first to discover cross-modal causal structures for the event-level visual question answering task.
\item We introduce a linguistic back-door causal intervention module guided by linguistic semantic relations to mitigate the spurious biases and uncover the causal dependencies within the  linguistic modality. To disentangle the visual spurious correlations, we propose a \textbf{L}ocal-\textbf{G}lobal \textbf{C}ausal \textbf{A}ttention \textbf{M}odule (LGCAM) that aggregates the local and global visual representations by front-door causal intervention.
\item We construct a \textbf{S}patial-\textbf{T}emporal \textbf{T}ransformer (STT) that models the multi-modal co-occurrence interactions between the visual and linguistic knowledge, to discover the fine-grained interactions among linguistic semantics, spatial, and temporal representations.
\item To adaptively fuse the causality-aware visual and linguistic features, we introduce a \textbf{V}isual-\textbf{L}inguistic \textbf{F}eature \textbf{F}usion (VLFF) module that leverages the hierarchical linguistic semantic relations to learn the global semantic-aware visual-linguistic features.
\item Extensive experiments on SUTD-TrafficQA, TGIF-QA, MSVD-QA, and MSRVTT-QA datasets show the effectiveness of our CMCIR for discovering visual-linguistic causal structures and achieving promising event-level visual question answering performance.
\end{itemize}
}
\section{Related Works}
\subsection{Visual Question Answering}
Compared \red{to} image-based visual question answering (i.e., ImageQA) \cite{antol2015vqa,yang2016stacked,anderson2018bottom}, event-level visual question answering (i.e., VideoQA) is much more challenging due to the extra temporal dimension. To \red{solve} the VideoQA problem, the model needs to capture spatial-temporal and visual-linguistic relations to infer the answer. To explore relational reasoning in VideoQA, Xu et al. \cite{xu2017video} proposed an attention mechanism to exploit the appearance and motion knowledge with the question as a guidance. Jang et al. \cite{jang2017tgif,jang2019video} released a large-scale VideoQA dataset named TGIF-QA and proposed a dual-LSTM based method with both spatial and temporal attention. Later on, some hierarchical attention and co-attention based methods \cite{li2019beyond,fan2019heterogeneous,jiayincai2020feature} are proposed to learn appearance-motion and question-related multi-modal interactions. Le et al. \cite{le2020hierarchical} proposed hierarchical conditional relation network (HCRN) to construct sophisticated structures for representation and reasoning over videos. Jiang et al. \cite{jiang2020reasoning} introduced the heterogeneous graph alignment (HGA) nework that aligns the inter- and intra-modality information for cross-modal reasoning. Huang et al. \cite{huang2020location} proposed \red{a} location-aware graph convolutional network to reason over detected objects. Lei et al. \cite{lei2021less} employed sparse sampling to build a transformer-based model named CLIPBERT and achieve end-to-end video-and-language understanding. Liu et al. \cite{liu2021hair} proposed a hierarchical visual-semantic relational reasoning (HAIR) framework to perform hierarchical relational reasoning.

Unlike \red{the} works that focus on relatively simple events like movie, TV-show or synthetic videos, our CMCIR framework \red{focuses} on complex event-level visual question answering and performs cross-modal causal relational reasoning \red{on} the spatial-temporal and linguistic content.  The only existing work for event-level urban visual question answering is Eclipse \cite{xu2021sutd}, which built an event-level urban traffic visual question answering dataset and proposed an efficient glimpse network to achieve computation-efficient and reliable video reasoning. Different from the Eclipse that focuses on the exploration of the efficient and dynamic reasoning in urban traffic events, our work aims to uncover the causal structures behind the visual-linguistic modalities and models the interaction between the appearance-motion and language knowledge in a causality-aware manner. In addition, these previous works tend to capture spurious linguistic or visual correlations within the videos, while we build a Causality-aware Visual-Linguistic Reasoning (CVLR) module to mitigate the bias caused by confounders and uncover the causal structures for the integration of complex event-level visual and linguistic modalities.

\begin{figure*}[ht]
\begin{center}
\includegraphics[scale=0.35]{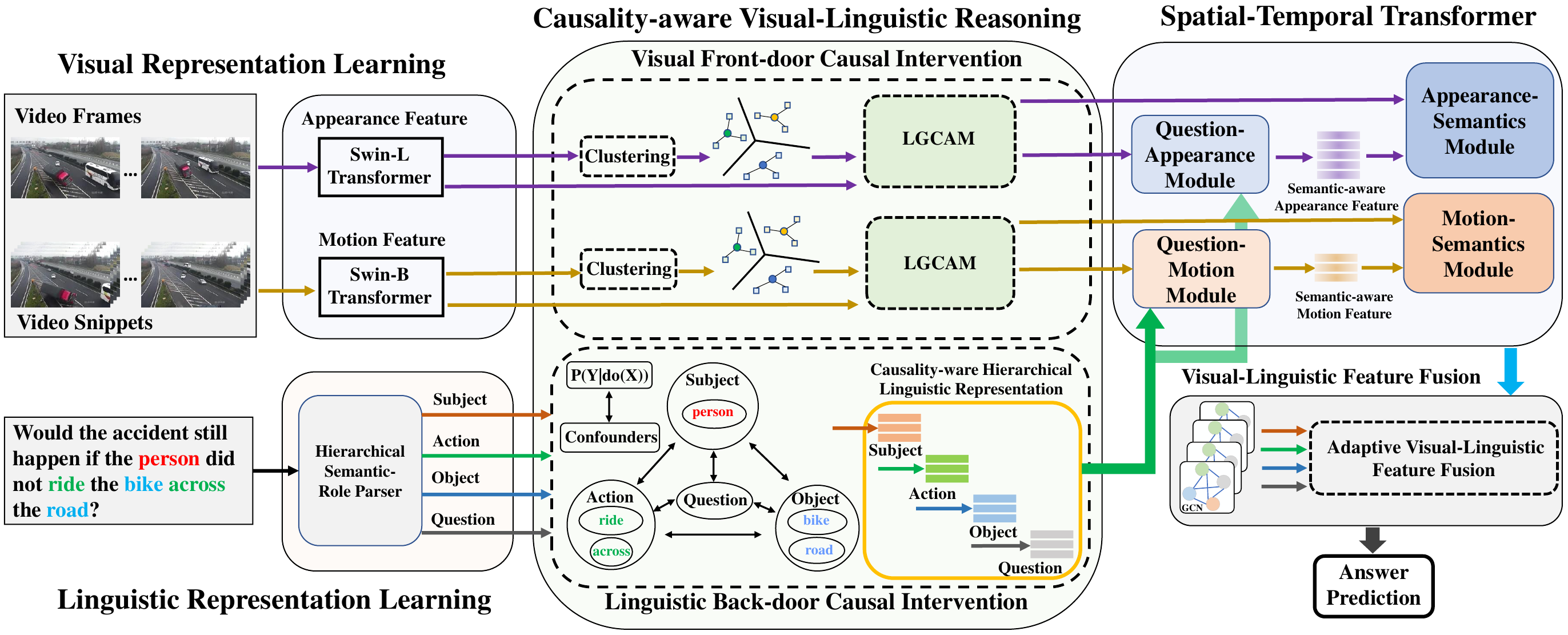}
\end{center}
   \vspace{-15pt}
   \caption{\red{Overview of CMCIR. The Linguistic Representation Learning (LRL) aims to parse the question into relation-centered tuples (subject, action, object) and then learns the hierarchical linguistic representations. The Causality-aware Visual-Linguistic Reasoning (CVLR) contains a visual front-door causal intervention module and a linguistic back-door causal intervention module. The visual front-door causal intervention module contains the Local-Global Causal Attention Module (LGCAM) that aggregates the local and global appearance and motion representations in a causality-aware way. The linguistic back-door causal intervention module models the linguistic confounder set from the perspective of semantic roles and de-confounds the language bias based on a structured causal model (SCM). Based on the causality-aware visual and linguistic representations, the Spatial-Temporal Transformer (STT) models the interaction between the appearance-motion and language knowledge in a coarse-to-fine manner. Finally, the Visual-Linguistic Feature Fusion (VLFF) module applies semantic graph guided adaptive feature fusion to obtain the multi-modal output.}}
      \vspace{-10pt}
\label{Fig3}
\end{figure*}

\subsection{Relational Reasoning for Event Understanding}
Besides VideoQA, relational reasoning has been explored in other event understanding tasks, such as action recognition  \cite{liu2018global,liu2018hierarchically,liu2019deep} and spatial-temporal grounding \cite{zhu2022hybrid}. To recognize and localize actions, Girdhar et al. \cite{girdhar2019video} introduced a transformer-style architecture to aggregate features from the spatiotemporal context around the person. For action detection, Huang et al. \cite{huang2019dynamic} introduced a dynamic graph module to model object-object interactions in video actions. Ma et al. \cite{ma2018attend} utilized an LSTM to model interactions between arbitrary subgroups of objects. Mavroudi et al. \cite{mavroudi2020representation} built a symbolic graph using action categories. Pan et al. \cite{pan2021actor} designed a high-order actor-context-actor relation network to realize indirect relation reasoning for spatial-temporal action localization. To localize a moment from videos for a given textual query, Nan et al. \cite{nan2021interventional} introduced a dual contrastive learning approach to align the text and video by maximizing the mutual information between semantics and video clips. Wang et al. \cite{wang2022weakly} proposed a causal framework to learn the deconfounded object-relevant association for robust video object grounding. However, these methods only perform relational reasoning over visual modality and neglects \red{potential causal} structures from linguistic semantic relation, resulting in incomplete and unreliable understanding of visual-linguistic content. Additionally, our CMCIR conducts causality-aware spatial-temporal relational reasoning to uncover the causal structure for visual-linguistic modality \red{and} utilizes hierarchical semantic knowledge for spatial-temporal relational reasoning.

\subsection{Causal Inference in Visual Representation Learning}
Compared to the conventional debiasing techniques \cite{wang2020devil}, causal inference \cite{pearl2016causal,yang2021deconfounded,liu2022causal} shows its potential in mitigating spurious correlations \cite{bareinboim2012controlling} and disentangling model effects \cite{besserve2020counterfactuals} for better generalization. Counterfactual and causal inference have attracted increasing attention in several computer vision tasks, including visual explanations \cite{goyal2019counterfactual,wang2020scout}, scene graph generation \cite{chen2019counterfactual,tang2020unbiased}, image recognition \cite{wang2020visual,wang2021causal}, video analysis \cite{fang2019modularized,kanehira2019multimodal,nan2021interventional}, and vision-language tasks \cite{abbasnejad2020counterfactual,niu2021counterfactual,yang2021causal,liu2023causality,chen2023visual,wei2023visual}. Specifically, Tang et al. \cite{tang2020long}, Zhang et al. \cite{zhang2020causal}, Wang et al. \cite{wang2020visual}, and Qi et al. \cite{qi2020two} computed the direct causal effect and \red{mitigated} the bias based on observable confounders. Counterfactual based solutions are also effective, for example, Agarwal et al. \cite{agarwal2020towards} proposed a counterfactual sample synthesising method based on GAN \cite{goodfellow2014generative}. Chen et al. \cite{chen2020counterfactual} tried to replace critical objects and critical words with \red{a} mask token and reassigned a answer to synthesis counterfactual QA pairs. Apart from sample synthesising, Niu et al. \cite{niu2021counterfactual} developed a counterfactual VQA framework that reduce multi modality bias by using causality approach named Natural Indirect Effect and Total Direct Effect to eliminate the mediator effect. Li et al. \cite{li2022invariant} proposed an Invariant Grounding for VideoQA (IGV) to force the VideoQA models to shield the answering process from the negative influence of spurious correlations.  Liu et al. \cite{liu2023causality}  introduced Visual Causality Discovery (VCD) architecture to find question-critical scene temporally and
disentangle the visual spurious correlations by the front-door causal intervention.

However, most of the existing causal visual tasks are relatively simple without considering more challenging tasks such as video understanding and event-level visual question answering. Although some recent works CVL \cite{abbasnejad2020counterfactual}, Counterfactual VQA \cite{niu2021counterfactual}, CATT \cite{yang2021causal}, IGV \cite{li2022invariant} and VCD \cite{liu2023causality} focused on visual question answering tasks, they adopted structured causal model (SCM) to eliminate either the linguistic or visual bias without considering cross-modal causality discovery.  Different from previous methods,  our CMCIR aims for event-level visual question answering that requires fine-grained understanding of spatial-temporal visual relation, linguistic semantic relation, and visual-linguistic causal dependency. Moreover, our Causality-aware Visual-Linguistic Reasoning (CVLR) applies front-door and back-door causal intervention modules to discover cross-modal causal structures.

\section{Methodology}
The framework of the CMCIR is shown in Fig. \ref{Fig3}, which is an event-level visual question answering architecture. In this section, we present the detailed implementations of  CMCIR.

\begin{figure}
\begin{center}
\includegraphics[scale=0.4]{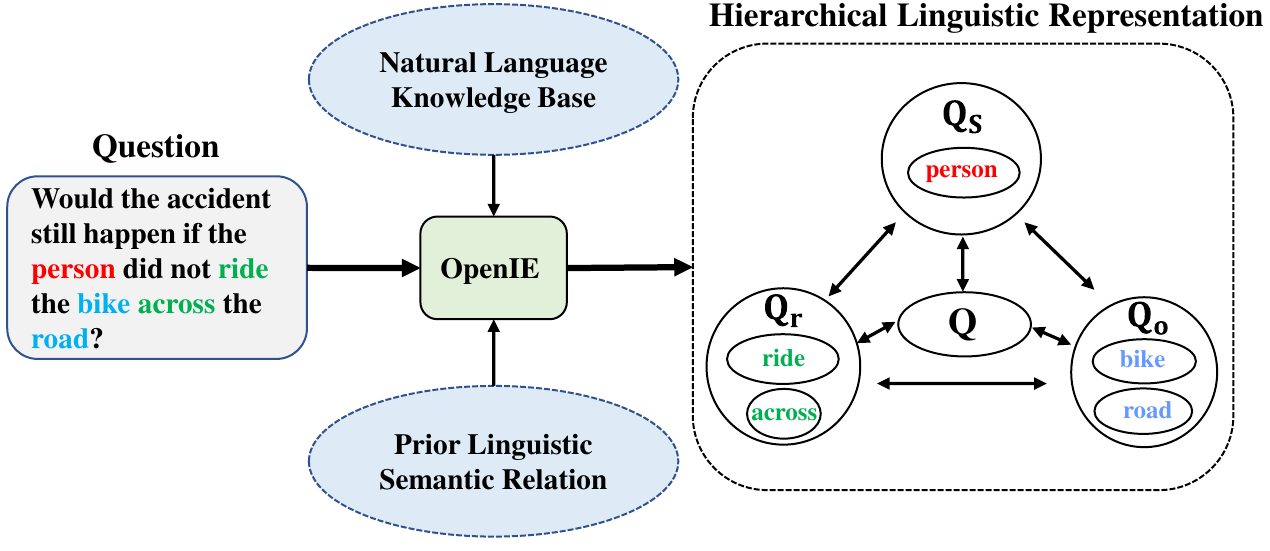}
\end{center}
   \vspace{-15pt}
   \caption{The proposed Hierarchical Semantic-Role Parser (HSRP) parses the question into verb-centered relation tuples (subject, action, object).}
      \vspace{-10pt}
\label{Fig4}
\end{figure}

\begin{figure*}
\begin{center}
\includegraphics[scale=0.235]{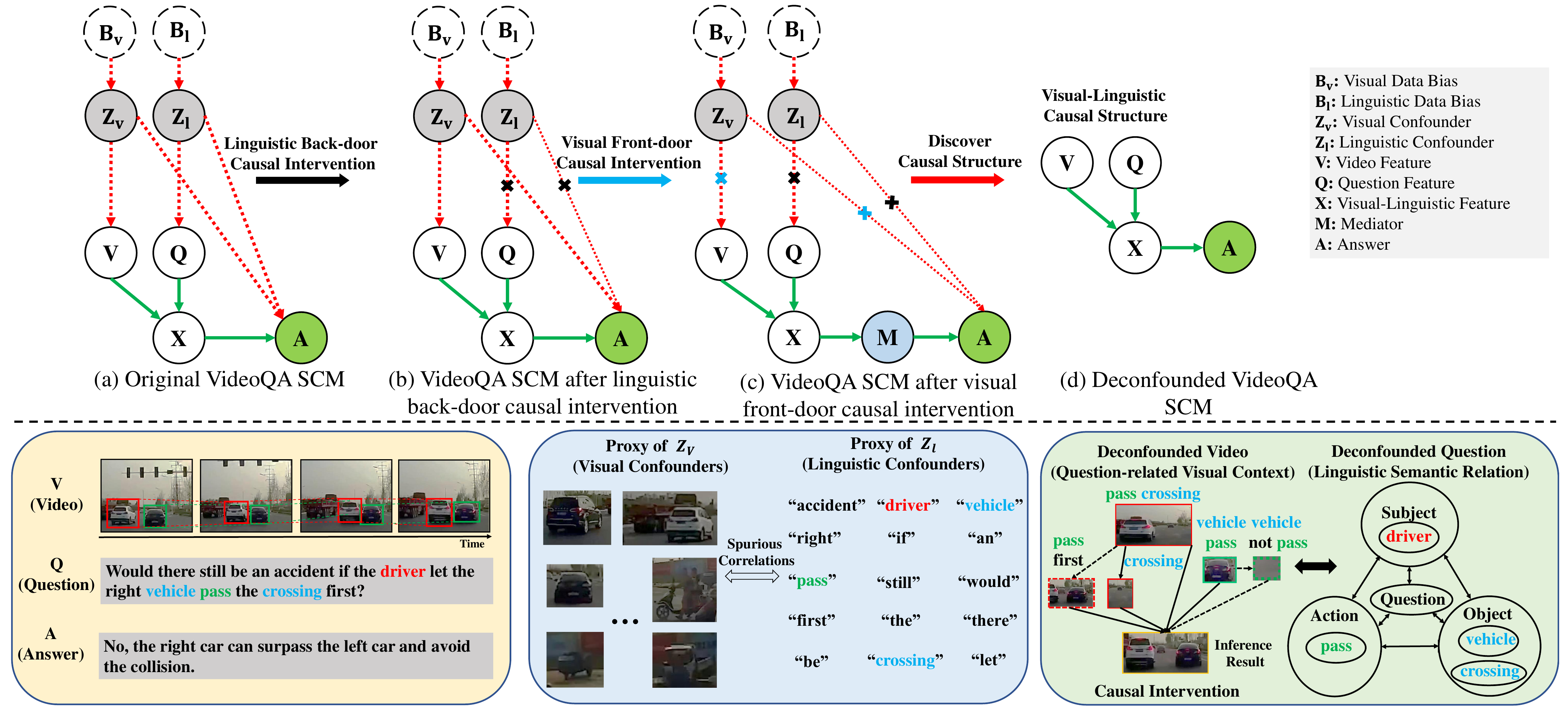}
\end{center}
   \vspace{-15pt}
   \caption{\red{The proposed causal graph of visual-linguistic causal intervention. The green path represents the unbiased visual question answering, which is the true causal effect. The red path shows the biased visual question answering caused by the confounders, also known as the back-door path. The bottom part of the figure provides an intuitive explanation of a real VideoQA sample using visual-linguistic causal intervention.}}
      \vspace{-10pt}
\label{Fig5}
\end{figure*}

\subsection{Visual Representation Learning}
The goal of event-level visual question answering is to deduce an answer $\tilde{a}$ from a video $\mathcal{V}$ with a given question $q$. The answer $\tilde{a}$ can be found in an answer space $\mathcal{A}$ which is a \red{predefined} set of possible answers for open-ended questions or a list of answer candidates for \red{multiple-choice} questions. The video $\mathcal{V}$ of $L$ frames is divided into $N$ equal clips. Each clip of $C_i$ of length $T=\lfloor L/N\rfloor$ is presented by two types of visual \red{feature}: frame-wise appearance feature vectors $F^a_i=\{f_{i,j}^a|f_{i,j}^a\in \mathbb{R}^{1536}, j=1,\ldots,T\}$ and motion feature vector at clip level $f^m_i\in \mathbb{R}^{1024}$. In our experiments, Swin-L \cite{liu2021swin} is used to extract the frame-level appearance features $F^a$ and Video Swin-B \cite{Liu_2022_CVPR} is applied to extract the clip-level motion features $F^m$. Then, we use a linear feature transformation layer to map $F^a$ and $F^m$ into the same $d$-dimensional feature space. Thus, we have $f_{i,j}^a, f^m_i\in \mathbb{R}^{d}$.

\subsection{Linguistic Representation Learning}\label{sec3.3}
From the perspective of linguistic semantic relations, a question usually contains the vocabulary of \red{a} subject, \red{an} action, and \red{an} object, since most videos can be described as ``somebody \red{doing} something''.  Therefore, we propose an efficient approach to approximate the confounder set distribution from the perspective of natural language. Specifically, we build a Hierarchical Semantic-Role Parser (HSRP) to parse the question into verb-centered relation tuples (subject, action, object) and construct three sets of vocabulary accordingly. The verb-centered relation tuples are subsets of the words of the original question around the key words subject, action, and object. The HSRP is based on the state-of-the-art Open Information Extraction (OpenIE) model \cite{stanovsky2018supervised}, which discovers linguistic semantic relations from \red{a} large-scale natural language knowledge base, as shown in Fig. \ref{Fig4}.  For the whole question $Q$, subject $Q_s$, action $Q_r$, object $Q_o$, and answer candidates $A$, each word is respectively embedded into a vector of $300$ \red{dimensions} by adopting pre-trained GloVe \cite{pennington2014glove} word embedding, which is further mapped into a $d$-dimensional space using linear transformation. Then, we represent the corresponding question and answer semantics as $Q=\{q_1,q_2,\cdots, q_L\}$, $Q_s=\{q^s_1,q^s_2,\cdots, q^s_{L_s}\}$, $Q_r=\{q^r_1,q^r_2,\cdots, q^r_{L_r}\}$, $Q_o=\{q^o_1,q^o_2,\cdots, q^o_{L_o}\}$, $A=\{a_1,a_2,\cdots, a_{L_a}\}$, where $L$, $L_s$, $L_r$, $L_o$, $L_a$ indicate the length of $Q$, $Q_s$, $Q_r$, $Q_o$, and $A$.

To obtain contextual linguistic representations that aggregate dynamic long-range temporal dependencies from multiple time-steps, a BERT \cite{devlin2018bert} model is employed to encode $Q$, $Q_s$, $Q_r$, $Q_o$, and the answer $A$, respectively. Finally, the updated representations for the question, question tuples, and answer candidates can be written as:
\begin{equation}\label{eq2}
\begin{aligned}
&Q=\{q_i|q_i\in \mathbb{R}^{d}\}_{i=1}^{L}, ~~Q_s=\{q^s_i|q^s_i\in \mathbb{R}^{d}\}_{i=1}^{L_s},\\
&Q_r=\{q^r_i|q^r_i\in \mathbb{R}^{d}\}_{i=1}^{L_r}, ~~Q_o=\{q^o_i|q^o_i\in \mathbb{R}^{d}\}_{i=1}^{L_o}
\end{aligned}
\end{equation}
and
\begin{equation}\label{eq3}
A=\{a_i|a_i\in \mathbb{R}^{d}\}_{i=1}^{L_a}
\end{equation}


\subsection{Causality-aware Visual-Linguistic Reasoning}
For visual-linguistic question reasoning with spatial-temporal data, we employ Pearl's structural causal model (SCM) \cite{pearl2016causal} to model the causal effect between video-question pairs and the answer, as shown in Fig. \ref{Fig5} (a). The nodes are variables and edges are causal relations. Conventional VQA methods only learn: $\{V,Q\} \rightarrow X\rightarrow A$, which learn the ambiguous statistics-based association $P(A|V,Q)$. They ignore the spurious association brought by the confounder, while our method \red{address} these problems in a causal \red{framework} and propose a fundamental solution. In the following, we detail the rationale behind our causal graph. The bottom part of Fig. \ref{Fig5} presents the high-level explanation of the visual-linguistic causal intervention. Here, we \red{provide} the detailed interpretation for some subgraphs.

$\{B_v,B_l\}\rightarrow \{Z_v,Z_l\}\rightarrow \{V, Q\}$. The visual  and linguistic confounders $Z_v$ and $Z_l$ (probably \red{an} imbalanced distribution of \red{the} dataset caused by data sampling biases $B_v$ and $B_l$) may lead to spurious correlations between videos and certain words. The $do$-operation on $\{V,Q\}$ can enforce their values and cuts off the direct dependency between  $\{V,Q\}$ and their parents $Z_v$ and $Z_l$ (Fig. \ref{Fig5} (b) and (c)).

$\{B_v,B_l\}\rightarrow \{Z_v,Z_l\}\rightarrow A$. Since $Z_v$ and $Z_l$ are the visual  and linguistic confounders for the dataset, we must also have $Z_v$ and $Z_l$ connected to prediction $A$ via directed paths excluding $\{V,Q\}$. This ensures the consideration of \red{the} confounding impact from $Z_v$ and $Z_l$ to $A$.

$A\leftarrow \{Z_v,Z_l\} \rightarrow \{V, Q\}\rightarrow X$.  There are two back-door paths where confounders $Z_v$ and $Z_l$ affect the video $V$ and question $Q$ respectively, and \red{ultimately} affect answer $A$, leading the model to learn the spurious association. As discussed before, if we had successfully cut off the path $\{Z_v,Z_l\} \nrightarrow \{V, Q\}\rightarrow X\rightarrow A$, $\{V,Q\}$ and $A$ are deconfounded and the model can learn the true causal effect $\{V,Q\}\rightarrow X\rightarrow A$.

To train a video question answering model that learns the true causal effect $\{V,Q\}\rightarrow X\rightarrow A$: the model should reason the answer $A$ from \red{the} video $V$ and \red{the} question $Q$ instead of exploiting the spurious correlations induced by the confounders $Z_v$ and $Z_l$ (i.e., overexploiting the co-occurrence between the visual and linguistic concepts).  For example, since the answer to the question ``What the color of the vehicle involved in the accident?'' is ``white'' in most cases, the model will easily learn the spurious correlation between the  concepts ``vehicle'' and ``white''. Conventional visual-linguistic question reasoning models usually focus on correlations between video and question by directly learning $P(A|V,Q)$ without considering the confounders $Z_v$ and $Z_l$. Thus, when given an accident video of black vehicle, the model still predicts answer ``white'' with strong confidence. In our SCM, the non-interventional prediction can be expressed using Bayes rule as:
\begin{equation}\label{eq4}
P(A|V,Q)=\sum_zP(A|V,Q,z)P(z|V,Q)
\end{equation}
However, the above objective learns not only the main direct correlation from $\{V,Q\}\rightarrow X\rightarrow A$ but also the spurious one from the unblocked back-door path $\{V,Q\}\leftarrow Z\rightarrow A$. An intervention on $\{V,Q\}$ is denoted as $do(V,Q)$, which cuts off the link $\{V,Q\}\leftarrow Z$ to block the back-door path $\{V,Q\}\leftarrow Z\rightarrow A$ and the spurious correlation is eliminated. In this way, $\{V,Q\}$ and $A$ are deconfounded and the model can learn the true causal effect $\{V,Q\}\rightarrow X\rightarrow A$. Actually, there are two techniques to calculate $P(A|do(V,Q))$, which are the back-door and front-door adjustments \cite{pearl2016causal,pearl2018book}, respectively. The back-door adjustment is effective when the confounder is observable. However, for the visual-linguistic question reasoning, the confounder in visual and linguistic modalities \red{is} not always observable. Thus, we propose both back-door and front-door causal intervention modules to discover the causal structure and disentangle the linguistic and visual biases based on their characteristics.

\subsubsection{Linguistic Back-door Causal Intervention}
For linguistic modality, the confounder set $Z_l$ caused by selection bias cannot be observed directly due to the unavailability of the sampling process. Due to the existence of linguistic confounders, existing approaches that mainly rely on the entire question representations \red{tend} to capture spurious linguistic correlations and \red{ignore} semantic roles embedded in questions. To mitigate the bias caused by confounders and uncover the causal structure behind the linguistic modality, we design a back-door adjustment strategy that approximates the confounder set distribution from
the perspective of linguistic semantic relations. Based on the linguistic representation learning in Section \ref{sec3.3}, our latent confounder set is approximated based on the verb-centered relation roles for the whole question, subject-related question, action-related question, object-related question $Q$, $Q_s$, $Q_r$, $Q_o$. Blocking the back-door path $B_l\rightarrow Z_l \rightarrow Q$ makes $Q$ have a fair opportunity to incorporate causality-aware factors for prediction (as shown in Fig. \ref{Fig5} (b)). The back-door adjustment calculates the interventional distribution $P(A|V,do(Q))$:
\begin{equation}\label{eq5}
\begin{aligned}
P(A|V,do(Q))&=\sum_{z_l}P(A|V,do(Q),z_l)P(z_l|V,do(Q))\\
&\approx\sum_{z_l}P(A|V,do(Q),{z_l})P({z_l})
\end{aligned}
\end{equation}

To implement the theoretical and imaginative intervention in Eq. (\ref{eq5}), we approximate the confounder set $Z_l$ to a set of verb-centered relation vocabularies $Z_l=[z_1,z_2,z_3,z_4]=[Q,Q_s,Q_r,Q_o]$. We compute the prior probability $P({z_l})$ in Eq. (\ref{eq5}) for verb-centered relation phrases $z$ in each set $z_1$, $z_2$, $z_3$, $z_4$ based on the dataset statistics:
\begin{equation}\label{eq6}
\begin{aligned}
P(z_l)=\frac{|z_l|}{\sum_{j\in z_l^i}|j|},~\forall z_l\in z_l^i, ~i=1,\cdots,4
\end{aligned}
\end{equation}
where $z_l^i$ is one of the four verb-centered relation vocabulary sets, $|z_l|$ is the number of samples in $z_l$, and $|j|$ is the number of occurrences of the phrase $j$. The representation of $z_l$ is calculated \red{in a similar way as in} Eq. (\ref{eq2}). Since $P(A|V,do(Q))$ is calculated by softmax, we apply Normalized Weighted Geometric Mean (NWGM) \cite{xu2015show} to Eq. (\ref{eq5}) to approximate the deconfounded prediction:
\begin{equation}\label{eq7}
\begin{aligned}
P(A|V,do(Q))&=\sum_{z_l}P(A|V,\textrm{concat}(Q, {z_l}))P({z_l})\\
&\approx P(A|\sum_{z_l}(V,\textrm{concat}(Q, {z_l}))P({z_l}))
\end{aligned}
\end{equation}
where $\textrm{concat}(\cdot)$ \red{represents} vector concatenation. According to Eq. (\ref{eq7}), each \red{element} of the causality-aware hierarchical
linguistic representation $Q^h=\{Q,Q_s,Q_r,Q_o\}$  \red{needs} to be integrated into the QA inference phase \red{using} Eq. (\ref{eq7}), which is essentially a weighted sum of the occurrences of the values of the linguistic \red{confounders} in the dataset.

\subsubsection{Visual Front-door Causal Intervention}
As shown in Eq. (\ref{eq5}), the back-door adjustment requires us to determine what the confounder is in advance. However, in visual domains, data biases are complex and it is hard to know and disentangle different types of confounders. Existing approaches usually define the confounders as the average of visual features \cite{wang2020visual,wang2021causal}. Actually, the average features may not properly \red{describe} a certain confounder especially for complex heterogeneous spatial-temporal data. Fortunately, the front-door adjustment \red{gives} a feasible way to calculate $P(A|do(V),Q)$ when we cannot explicitly represent the confounder. As shown in Fig. \ref{Fig5} (c), to apply the front-door adjustment, an additional mediator $M$ should be inserted between $X$ and $A$ to construct a front-door path $V\rightarrow X\rightarrow M\rightarrow A$ \red{to transmit} knowledge. For visual-linguistic question reasoning task, an attention-based model will select a few regions from the video $V$  based on the question $Q$ to predict the answer $A$, where $m$ denotes the selected knowledge from mediator $M$:
\begin{equation}\label{eq8}
\begin{aligned}
P(A|V,Q)=\sum_mP(M=m|V,Q)P(A|M=m)
\end{aligned}
\end{equation}

\begin{figure}
\begin{center}
\includegraphics[scale=0.56]{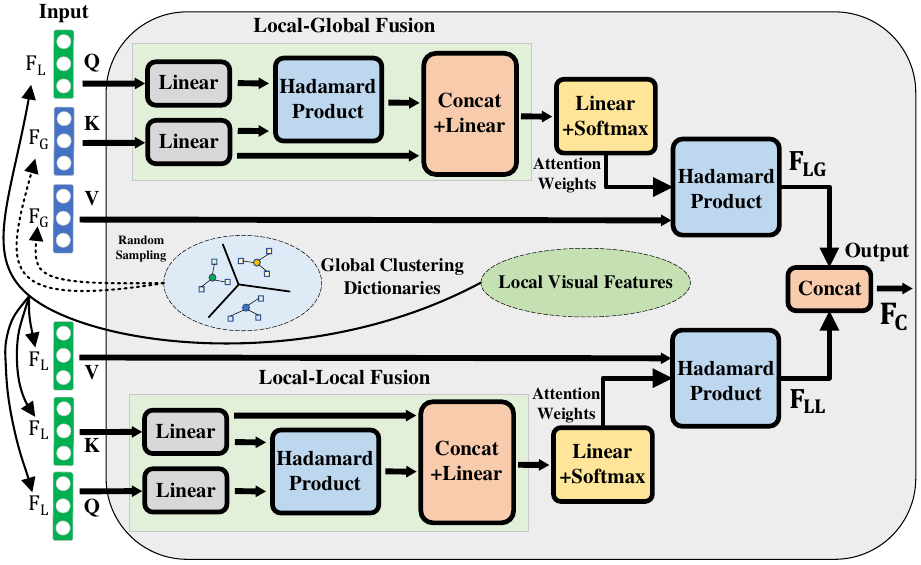}
\end{center}
   \vspace{-15pt}
   \caption{\red{The structure of Local-Global Causal Attention Module (LGCAM), which jointly estimates $\hat{M}$ and $\hat{V}$ in an unified attention module.}  }
      \vspace{-15pt}
\label{Fig6}
\end{figure}

Then, the answer predictor can be represented by two parts: a feature extractor $V\rightarrow X\rightarrow M$ and a answer predictor $M\rightarrow A$. Thus, the interventional probability $P(A|do(V),Q)$ can be represented as:
\begin{equation}\label{eq9}
\begin{aligned}
P(A|do(V),Q)&=\sum_mP(M=m|do(V),Q)P(A|do(M=m))
\end{aligned}
\end{equation}

Next, we discuss the above feature extractor $V\rightarrow X\rightarrow M$ and answer predictor $M\rightarrow A$, respectively.

\textbf{Feature Extractor $V\rightarrow X\rightarrow M$}. As shown in Fig. \ref{Fig5} (c), for the causal link $V\rightarrow X\rightarrow M$, the back-door path between $V$ and $M$: $X\leftarrow V\leftarrow Z_v\rightarrow M\rightarrow A$ is already blocked. Thus, the interventional probability is equal to the conditional one:
\begin{equation}\label{eq10}
\begin{aligned}
P(M=m|do(V),Q)=P(M=m|V,Q)
\end{aligned}
\end{equation}

\textbf{Answer Predictor  $M\rightarrow A$}. To realize $P(A|do(M=m))$, we can cut off $M\leftarrow X$ to block the back-door path $M\leftarrow X\leftarrow V\leftarrow Z_v\rightarrow A$:
\begin{equation}\label{eq11}
\begin{aligned}
P(A|do(M=m))=\sum_vP(V=v)P(A|V=v,M=m)
\end{aligned}
\end{equation}

To sum up, by applying Eq. (\ref{eq10}) and Eq. (\ref{eq11}) into Eq. (\ref{eq9}), we can calculate the true causal effect between $V$ and $A$:
\begin{equation}\label{eq12}
\begin{aligned}
&P(A|do(V),Q)=\\
&\sum_mP(M=m|V,Q)\sum_vP(V=v)P(A|V=v,M=m)
\end{aligned}
\end{equation}

To implement visual front-door causal intervention Eq. (\ref{eq12}) in a deep learning framework, we parameterize the $P(A|V,M)$ as a network $g(\cdot)$ followed by a softmax layer since most of visual-linguistic tasks are transformed into classification formulations:
\begin{equation}\label{eq13}
\begin{aligned}
P(A|V,M)=\textrm{Softmax}[g(M,V)]
\end{aligned}
\end{equation}

From Eq. (\ref{eq12}),  we can see that both $V$ and $M$ are required to be sampled and fed into the network to complete $P(A|do(V),Q)$. However, the cost of forwarding all the samples is \red{high}. To \red{tackle} this problem, we apply \red{the} Normalized Weighted Geometric Mean (NWGM) \cite{xu2015show} to \red{incorporate} the outer sampling into the feature level, \red{thereby requiring only one forward pass of the} ``absorbed input" in the network, as \red{shown} in Eq. (\ref{eq14}):
\begin{equation}\label{eq14}
\begin{aligned}
&P(A|do(V),Q)\approx \textrm{Softmax}[g(\hat{M},\hat{V})]\\
&=\textrm{Softmax}\Big[g(\sum_mP(M=m|f(V))m,\sum_vP(V=v|h(V))v)\Big]
\end{aligned}
\end{equation}
where $\hat{M}$ and $\hat{V}$ denote the estimations of $M$ and $V$, $h(\cdot)$ and $f(\cdot)$ denote the network mapping functions.

Actually, $\hat{M}$ is essentially an in-sample sampling process where $m$ denotes the selected knowledge from the current input sample $V$, \red{and} $\hat{V}$ is essentially a cross-sample sampling process since it comes from other samples. Therefore, both $\hat{M}$ and $\hat{V}$ can be calculated by attention networks \cite{yang2021causal}. Specifically, we propose a novel Local-Global Causal Attention Module (LGCAM) that jointly estimates $\hat{M}$ and $\hat{V}$ in \red{a} unified attention module to increase the representation ability of the causality-aware visual features. $\hat{M}$ can be calculated by learning local-local visual feature $F_{LL}$, $\hat{V}$ can be calculated by learning local-global visual feature $F_{LG}$. Here, we \red{use} the computation of $F_{LG}$ as \red{an} example to clarify our LGCAM, as shown in the upper part of Fig. \ref{Fig6}.

Specifically, we \red{first} calculate $F_L=f(V)$ and $F_G=h(V)$ and use them as the input of the LGCAM, where $f(\cdot)$ denotes the visual feature extractor (frame-wise appearance feature or motion feature) followed by a query embedding function, and $h(\cdot)$ denotes the \red{K-means-based} visual feature selector from the whole training samples followed by a query embedding function. Thus, $F_L$ represents the visual feature of the current input sample (local visual feature) and $F_G$ represents the global visual feature. The $F_G$ is obtained by randomly sampling from the whole clustering dictionaries with the same size as $F_L$. The LGCAM takes $F_L$ and $F_G$ as inputs and computes local-global visual feature $F_{LG}$ by conditioning global visual feature $F_G$ \red{on} the local visual feature $F_L$. The output of the LGCAM is denoted as $F_{LG}$, which is given by:
\begin{equation}\label{eq15}
\begin{aligned}
&\textbf{\textrm{Input}}: Q=F_L, K=F_G, V=F_G\\
&\textbf{\textrm{Local-Global Fusion}}: H=[W_VV,W_QQ\odot W_KK]\\
&\textbf{\textrm{Activation Mapping}}: H^\prime=\textrm{GELU}(W_HH+b_H)\\
&\textbf{\textrm{Attention Weights}}: \alpha=\textrm{Softmax}(W_{H^\prime}H^\prime+b_{H^\prime})\\
&\textbf{\textrm{Output}}: F_{LG}=\alpha\odot F_G
\end{aligned}
\end{equation}
where $[.,.]$ denotes \red{a} concatenation operation, $\odot$ \red{represents} the Hadamard product, $W_Q$, $W_K$, $W_V$, $W_{H^\prime}$ \red{represent} the weights of \red{the} linear layers,  $b_H$ and $b_{H^\prime}$  denote the biases of \red{the} linear layers. From Fig. \ref{Fig3}, the visual front-door causal intervention module has two branches for appearance and motion features. Therefore, the  $F_{LG}$ has two variants, \red{$F_{LG}^a$ for the appearance branch, and $F_{LG}^m$ for the motion branch}.

The $F_{LL}$ can be computed similarly as $F_{LG}$ when setting $Q=K=V=F_L$. Finally, the $F_{LG}$ and $F_{LL}$ are concatenated $F_C=[F_{LG},F_{LL}]$ for estimating $P(A|do(V),Q)$.

\begin{figure}
\begin{center}
\includegraphics[scale=0.31]{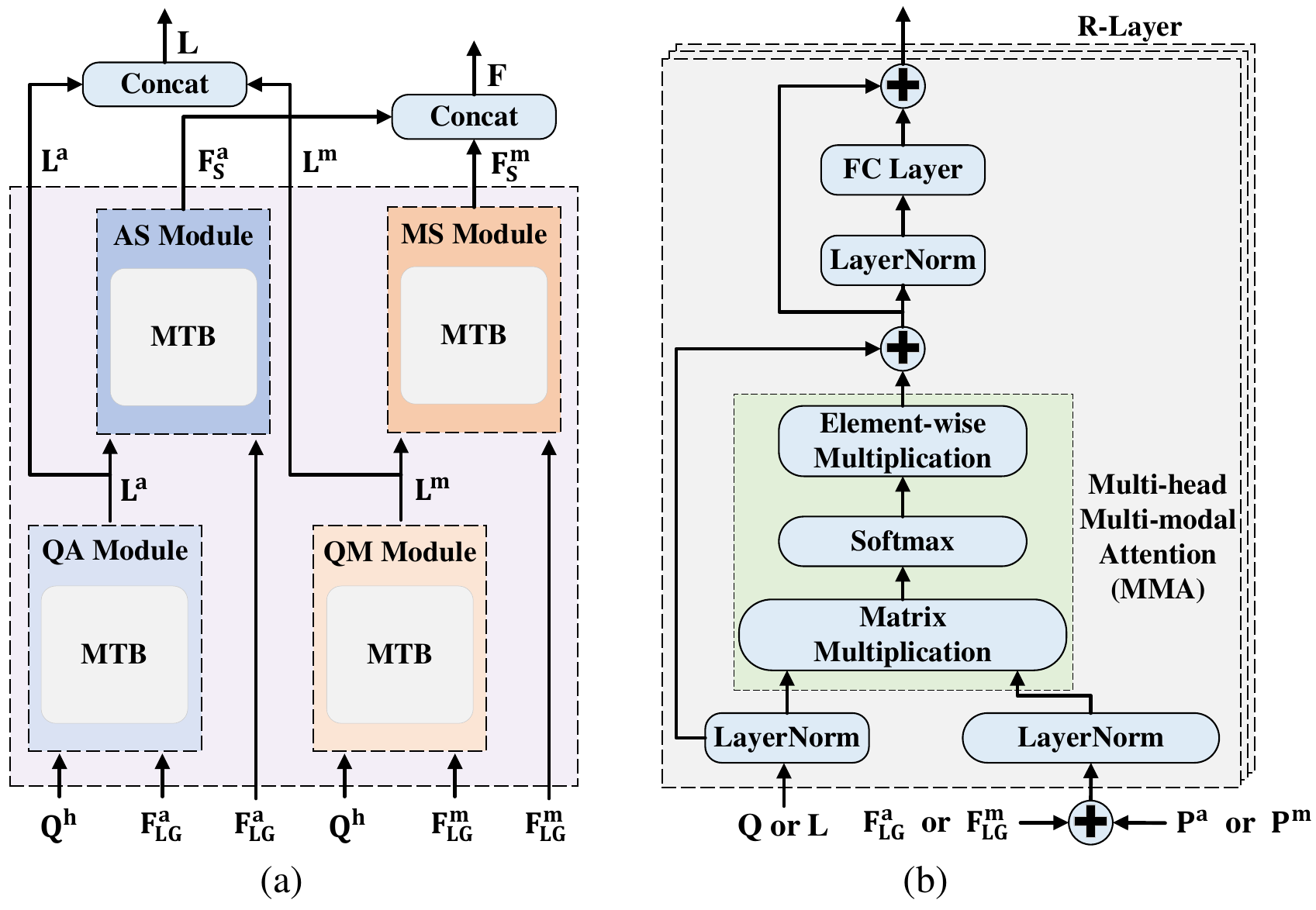}
\end{center}
   \vspace{-15pt}
   \caption{Illustration of the (a) Spatial-Temporal Transformer (STT), and the (b) Multi-modal Transformer Block (MTB) in the STT.  }
      \vspace{-15pt}
\label{Fig7}
\end{figure}

\subsection{Spatial-Temporal Transformer}
After performing linguistic and visual causal intervention, we need to conduct visual-linguistic relation modeling and feature fusion. However, existing vision-and-language transformers typically neglect the multi-level and fine-grained interaction between text and appearance-motion information, which is crucial for \red{the} event-level visual question answering task. Therefore, we propose a Spatial-Temporal Transformer (STT) that includes four sub-modules, namely Question-Appearance (QA), Question-Motion (QM), Appearance-Semantics (AS) and Motion-Semantics (MS), as depicted in Fig. \ref{Fig7} (a), to uncover the fine-grained interactions between linguistic and spatial-temporal representations. The QA (QM) module consists of an R-layer Multi-modal Transformer Block (MTB) (Fig. \ref{Fig7} (b)) for multi-modal interaction between the question and the appearance (motion) features. Similarly, the AS (MS) uses the MTB to deduce the appearance (motion) information given the question semantics.

The QA and AM modules aim to develop a comprehensive comprehension of the question concerning the visual appearance and motion content, respectively. For QA and QM modules, the input of MTB are $Q^h=\{Q,Q_s,Q_r,Q_o\}$ obtained from section 3.3.1 and $F_C^a$,  $F_C^m$ obtained from section 3.3.2,  respectively. To maintain the positional information of the video sequence, the appearance feature $F_C^a$ and motion feature $F_C^m$ are firstly added with the learned positional embeddings $P^a$ and $P^m$, respectively. Thus, for $r=1,2,\ldots, R$ layers of the MTB, with the input $F_C^a=[F_C^a,P^a]$, $F_C^m=[F_C^m,P^m]$, $Q^a$, and $Q^m$, the multi-modal output for QA and QM are computed as:
\begin{equation}\label{eq16}
\begin{aligned}
&\hat{Q}^a_r=U^a_r+\sigma^a(\textrm{LN}(U^a_r))\\
&\hat{Q}^m_r=U^m_r+\sigma^m(\textrm{LN}(U^m_r))\\
&U^a_r=\textrm{LN}(\hat{Q}^a_{r-1})+\textrm{MMA}^a(\hat{Q}^a_{r-1},F_C^a)\\
&U^m_r=\textrm{LN}(\hat{Q}^m_{r-1})+\textrm{MMA}^m(\hat{Q}^m_{r-1},F_C^m)\\
\end{aligned}
\end{equation}
where $\hat{Q}^a_0=Q^h, \hat{Q}^m_0=Q^h$, $U^a_r$ and $U^m_r$ are the intermediate \red{features} at \red{the} $r$-th layer of the MTB.  LN$(\cdot)$ denotes the layer normalization operation and $\sigma^a(\cdot)$  and $\sigma^m(\cdot)$ denote the linear projections. MMA$(\cdot)$ is the Multi-head Multi-modal Attention layer. We denote the output semantics-aware
appearance and motion features of QA and MA as $L^a=\hat{Q}^a=\hat{Q}^a_R$ and $L^m=\hat{Q}^m=\hat{Q}^m_R$, respectively.

Since an essential step of VideoQA is to infer the visual \red{information} within the appearance-motion \red{features} given the question semantics, we propose \red{the} Appearance-Semantics (AS) and Motion-Semantics (MS) modules to infer the visual information from the interactions between the linguistic semantics and the spatial-temporal representations, with \red{a} similar architecture \red{to the} Multi-modal Transformer Block (MTB). Given the semantics-aware appearance and motion features $L^a$ and $L^m$, we use \red{the} AS and MS to discover \red{visual information} to answer the question based on the spatial-temporal visual representations, respectively.

Similar to Eq. (\ref{eq16}), given the visual appearance and motion features $\hat{F}_{LG}^a$, $\hat{F}_{LG}^m$ and question semantics $L^a$, $L^m$, the multi-modal output for AS and MS are computed as:
\begin{equation}\label{eq19}
\begin{aligned}
&\hat{L}^a_r=U^a_r+\sigma^a(\textrm{LN}(U^a_r))\\
&\hat{L}^m_r=U^m_r+\sigma^m(\textrm{LN}(U^m_r))\\
&U^a_r=\textrm{LN}(F_{C, r-1}^a)+\textrm{MMA}^a(F_{C, r-1}^a, L^a)\\
&U^m_r=\textrm{LN}(F_{C, r-1}^m)+\textrm{MMA}^m(F_{C, r-1}^m, L^m)\\
\end{aligned}
\end{equation}
where the MTB has $r=1,2,\ldots,R$ layers, and $F_{C, 0}^a=F_{C}^a$, $F_{C, 0}^m=F_{C}^m$. The output visual clues of QA and MA are denoted as $F_{s}^a=\hat{L}^a_R$ and $F_{s}^m=\hat{L}^m_R$, respectively. Then, the output of the AS and MS \red{is} concatenated to make the final visual output $F=[F_{s}^a,F_{s}^m]\in \mathbb{R}^{2d}$. The output of the QA and QM are concatenated to make the  final question semantics output $L=[L^a,L^m]\in \mathbb{R}^{2d}$.

\begin{figure*}
\begin{center}
\includegraphics[scale=0.55]{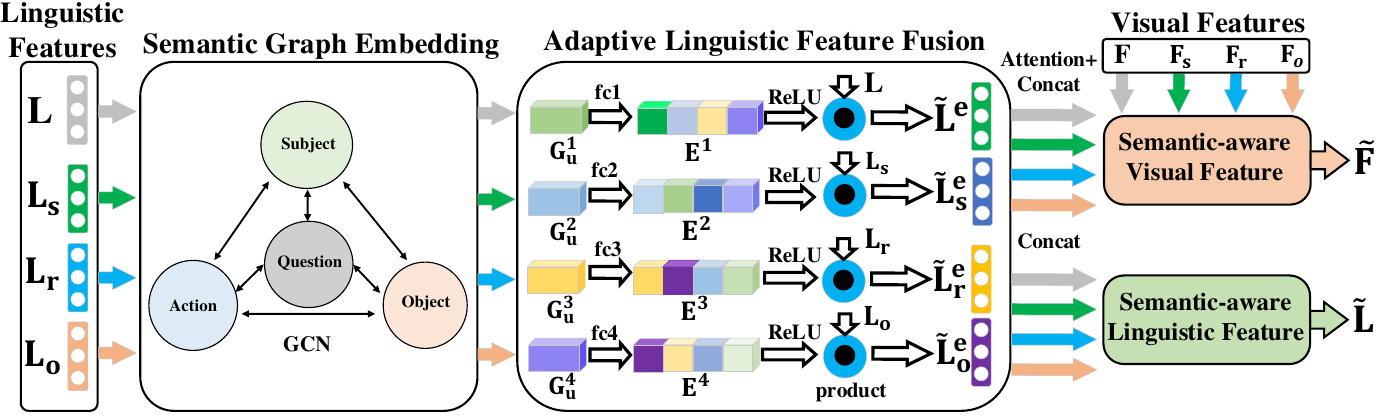}
\end{center}
   \vspace{-15pt}
   \caption{\red{Illustration of the Visual-Linguistic Feature Fusion (VLFF) module, which leverages the hierarchical linguistic semantic relations to learn the global semantic-aware visual-linguistic features, and finally fuses the causality-aware visual and linguistic features adaptively.} }
      \vspace{-10pt}
\label{Fig8}
\end{figure*}

\subsection{Visual-Linguistic Feature Fusion}
According to Eq. (\ref{eq7}) in section 3.4.1, each item of the causality-aware hierarchical linguistic representation $Q^h=\{Q,Q_s,Q_r,Q_o\}$  is required to conduct the QA prediction process respectively, and then integrate their results by their semantic relations. Thus, for $Q$, $Q_s$, $Q_r$, $Q_o$, their respective visual and linguistic outputs of the STT model are denoted as $F, F_s, F_r, F_o$ and $L, L_s, L_r, L_o$, respectively. Specifically, \red{a semantic graph is constructed},  and the representation of the graph nodes is denoted as $L_g=\{L, L_s, L_r, L_o\}$, as shown in Fig. \ref{Fig8}. The feature vectors in $L_g$ are treated as the nodes. According to the hierarchical linguistic semantic relations among $Q$, $Q_s$, $Q_r$ and $Q_o$ learned by the HSRP, we build the fully-connected edges and then perform $g$-layer semantic graph convolutional (GCN) \cite{kipf2016semi} embedding:
\begin{equation}\label{eq20}
\begin{aligned}
L_g^e=\textrm{GCN}(L_g)=\{L^e, L_s^e, L_r^e, L_o^e\}
\end{aligned}
\end{equation}
where $\textrm{GCN}(\cdot)$ denotes the $g$-layer graph convolutions.

\red{As the linguistic features from different semantic roles are correlated, we have built an adaptive linguistic feature fusion module that receives features from different semantic roles and learns a global context embedding. This embedding is then used to recalibrate the input features from different semantic roles, as shown in  Fig. \ref{Fig8}}. The linguistic features of nodes learned from semantic GCN are \red{denoted as} $\{L^e_1, L_2^e, L_3^e, L_4^e\}=\{L^e, L_s^e, L_r^e, L_o^e\}$, where $L^e_k\in \mathbb{R}^{2d} (k=1,\cdots,4$). \red{To leverage the correlation among linguistic features, we concatenate them and obtain joint representations} $G_u^k$ for each semantic role $L^e_k$ \red{by passing them through a fully-connected layer}:
\begin{equation}\label{eq21}
G_u^k=W_{s}^k[L^e_1, L_2^e, L_3^e, L_4^e]+b_{s}^k, ~~k=1,\cdots,4
\end{equation}
where $[\cdot,\cdot]$ denotes the concatenation operation, $G_u^k\in \mathbb{R}^{d_{u}}$ denotes the joint representation, $W_{s}^k$ and $b_{s}^k$ are weights and bias of the fully-connected layer. We choose $d_{u}=d$ to restrict the model capacity and increase its generalization ability. To \red{utilize} the global context information aggregated in the joint representations $G_{u}^k$, we predict \red{an} excitation signal for \red{them} via a fully-connected layer:
\begin{equation}\label{eq22}
E^k=W_{e}^kG_u^k+b_{e}^k, ~~k=1,\cdots,4
\end{equation}
where $W_{e}^k$ and $b_{e}^k$ are \red{the} weights and biases of the fully-connected layer. After obtaining the excitation signal $E^k\in \mathbb{R}^{c}$, we use it to \red{adaptively} recalibrate the input feature $L^e_k$ by a simple gating mechanism:
\begin{equation}\label{eq23}
\widetilde{L}^e_k=\delta(E^k)\odot L^e_k
\end{equation}
where $\odot$ is \red{a} channel-wise product operation for each element in the channel dimension, and $\delta(\cdot)$ is the ReLU function. In this way, we can allow the features of one semantic role to recalibrate the features of another semantic role while preserving the correlation among different semantic roles. Then, these refined linguistic feature vectors $\{\widetilde{L}^e, \widetilde{L}_s^e, \widetilde{L}_r^e, \widetilde{L}_o^e\}$ are concatenated to form the final semantic-ware linguistic feature $\widetilde{L}=[\widetilde{L}^e, \widetilde{L}_s^e,\widetilde{L}_r^e, \widetilde{L}_o^e]\in \mathbb{R}^{4d}$.

To obtain the semantic-aware visual feature, we compute the visual feature $\widetilde{F}_{k}$ by individually conditioning each semantic role from the visual features $\{F_1, F_2, F_3, F_4\}=\{F, F_s, F_r, F_o\}$ to each semantic role from the refined linguistic features $\{\widetilde{L}^e_1, \widetilde{L}_2^e, \widetilde{L}_3^e, \widetilde{L}_4^e\}=\{\widetilde{L}^e, \widetilde{L}_s^e, \widetilde{L}_r^e, \widetilde{L}_o^e\}$ using the same operation as \red{in} \cite{le2020hierarchical}. For each semantic role $k~(k=1,2,3,4)$, the weighted semantic-aware visual feature is:
\begin{equation}\label{eq24}
\begin{aligned}
&I_k=\textrm{ELU}\big(W_k^I[W_k^fF_k,W_k^fF_k\odot W_k^l\widetilde{L}^e_k]+b_k^I\big)\\
&\widetilde{F}_{k}=\textrm{Softmax}(W_k^{I^\prime}I_k+b_k^{I^\prime})\odot F_k
\end{aligned}
\end{equation}
Then, these semantic-aware visual features $\widetilde{F}_{k}~(k=1,\cdots,4)$ are concatenated to form the final semantic-aware visual feature $\widetilde{F}=[\widetilde{F}_{1},\widetilde{F}_{2},\widetilde{F}_{3},\widetilde{F}_{4}]\in \mathbb{R}^{4d}$. Finally,  we infer the answer based on the semantic-aware visual feature $\widetilde{F}$ and linguistic feature $\widetilde{L}$. Specifically, we apply different answer decoders \cite{le2020hierarchical} depending on the visual question reasoning tasks, which are divided into three types: open-ended, multi-choice, and counting.

\section{Experiments}
In this section, we conduct extensive experiments to evaluate the performance of our CMCIR model. To verify the effectiveness of CMCIR and its components, we compare CMCIR with state-of-the-art methods and conduct ablation studies. Then, we conduct parameter sensitivity analysis to evaluate how the hyper-parameters of CMCIR \red{affect} the performance. We further show some visualization analysis to validate the ability of causal reasoning of CMCIR.

\subsection{Datasets}
In this paper, we evaluate our CMCIR on \red{the} event-level urban dataset SUTD-TrafficQA \cite{xu2021sutd} and three benchmark real-world datasets TGIF-QA \cite{jang2017tgif}, MSVD-QA \cite{xu2017video}, and MSRVTT-QA \cite{xu2017video}. The detailed descriptions of these datasets are as follows:

\textbf{SUTD-TrafficQA}. This dataset consists of 62,535 QA pairs and 10,090 videos collected from traffic scenes. There are six challenging reasoning tasks including basic understanding, event forecasting, reverse reasoning, counterfactual inference, introspection and attribution analysis. The basic understanding task is to perceive and understand traffic scenarios at the basic level. The event forecasting task is to infer future events based on observed videos, and the forecasting questions query about the outcome of the current situation. The reverse reasoning task is to ask about the events that have happened before the start of a video. The counterfactual inference task queries the consequent outcomes of certain hypothesis that do not occur. The introspection task is to test if models can provide preventive advice that could have been taken to avoid traffic accidents. The attribution task seeks the explanation about the causes of traffic events and infer the underlying factors.

\begin{table}[t]\renewcommand\tabcolsep{4.0pt}\renewcommand\arraystretch{1.0}
\begin{center}
\begin{tabular}{lcccccc}
\hline
&Video&QA pairs&Count&Action&Transition&FrameQA\\\hline
Train &62,846&139,414&26,843&20,475&52,704&39,392\\
Test&9,575&25,751&3,554&2,274&6,232&13,691\\\hline
Total&71,741&165,165&30,397&22,749&58,936&53,083\\\hline
\end{tabular}
\end{center}
\caption{Statistics of the TGIF-QA dataset. }
\vspace{-20pt}
\label{Table1}
\end{table}

\begin{table}[t]\renewcommand\tabcolsep{4.0pt}\renewcommand\arraystretch{1.0}
\begin{center}
\begin{tabular}{lccccccc}
\hline
&Video&QA pairs&What&Who&How&When&Where\\\hline
Train &1,200&30,933&19,485&10,479&736&161&72\\
Val&250&6,415&3,995&2,168&185&51&16\\
Test&520&13,157&8,149&4,552&370&58&28\\\hline
Total&1,970&50,505&31,629&17,199&1,291&270&116\\\hline
\end{tabular}
\end{center}
\caption{Statistics of the MSVD-QA dataset. }
\vspace{-20pt}
\label{Table2}
\end{table}

\begin{table}[!t]\renewcommand\tabcolsep{4.0pt}\renewcommand\arraystretch{1.0}
\begin{center}
\begin{tabular}{lccccccc}
\hline
&Video&QA pairs&What&Who&How&When&Where\\\hline
Train &6,513&158,581&108,792&43,592&4,067&1,626&504\\
Val &497&12,278&8,337&3,439&344&106&52\\
Test&2,990&72,821&49,869&20,385&1,640&677&250\\\hline
Total&10,000&243,680&166,998&67,416&6,051&2,409&806\\\hline
\end{tabular}
\end{center}
\caption{Statistics of the MSRVTT-QA dataset. }
\vspace{-25pt}
\label{Table3}
\end{table}

\begin{table*}[t]\renewcommand\tabcolsep{8.0pt}\renewcommand\arraystretch{1}
\begin{center}
\begin{tabular}{lccccccc}
\hline
\multirow{3}*{Method}&\multicolumn{6}{c}{Question Type}\\\cline{2-8}
&Basic&Attribution&Introspection&Counterfactual&Forecasting&Reverse&All\\
&(4759)&(348)&(482)&(302)&(166)&(565)&(6622)\\\hline
VIS+LSTM \cite{ren2015exploring}&-&-&-&-&-&-&29.91\\
I3D+LSTM\cite{carreira2017quo}&-&-&-&-&-&-&33.21\\
BERT-VQA \cite{yang2020bert}&-&-&-&-&-&-&33.68\\
TVQA \cite{lei2018tvqa}&-&-&-&-&-&-&35.16\\
$\textrm{VQAC}^\dag$  \cite{kim2021video}&34.02&49.43&\underline{34.44}&39.74&38.55&49.73&36.00\\
$\textrm{MASN}^\dag$ \cite{seo2021attend}&33.83&\underline{50.86}&34.23&41.06&41.57&\underline{50.80}&36.03\\
$\textrm{DualVGR}^\dag$ \cite{wang2021dualvgr}&33.91&50.57&33.40&\underline{41.39}&41.57&50.62&36.07\\
HCRN \cite{le2020hierarchical}&-&-&-&-&-&-&36.49\\
$\textrm{HCRN}^\dag$ \cite{le2020hierarchical}&\underline{34.17}&50.29&33.40&40.73&\underline{44.58}&50.09&36.26\\
Eclipse \cite{xu2021sutd}&-&-&-&-&-&-&\underline{37.05}\\\hline
\textbf{CMCIR (ours)}&\textbf{36.10}  (+1.93)&\textbf{52.59}  (+1.73)&\textbf{38.38}  (+3.94)&\textbf{46.03} (+4.64)&\textbf{48.80 } (+4.22)&\textbf{52.21} (+1.41)&\textbf{38.58} (+1.53)\\\hline
\end{tabular}
\end{center}
\caption{Results on SUTD-TrafficQA dataset. `\dag' indicates the result re-implemented by the officially code. The \textbf{best} and \underline{second-best} results are highlighted.}
\vspace{-20pt}
\label{Table4}
\end{table*}

\textbf{TGIF-QA}. This dataset has 165K QA pairs collected from 72K animated GIFs. It has four tasks: repetition count, repeating action, state transition, and frame QA. Repetition count is a counting task that requires a model to count the number of repetitions of an action. Repetition action and state transition are \red{multiple-choice} tasks with $5$ optional answers. FrameQA is an open-ended task with a \red{predefined} answer set, which can be answered from a single video frame. Table \ref{Table1} shows the statistics of the TGIF-QA dataset.

\textbf{MSVD-QA}. This dataset is created from the Microsoft Research Video Description Corpus \cite{chen2011collecting}, \red{which} is widely used in the video captioning task. It consists of 50,505 algorithm-generated question-answer pairs and 1,970 trimmed video clips. Each video lasts approximately 10 seconds. \red{It contains five questions types:} What, Who, How, When, and Where. The statistics of the MSVD-QA dataset are presented in Table \ref{Table2}.

\textbf{MSRVTT-QA}. \red{This larger dataset contains} more complex scenes constructed from the MSRVTT \cite{xu2016msr}. It contains 10,000 trimmed video clips of approximately 15 seconds each. A total of 243,680 question-answer pairs contained in this dataset are automatically generated by the NLP algorithm. The dataset contains five question types: What, Who, How, When, and Where. The statistics of the MSRVTT-QA dataset are presented in Table \ref{Table3}.

\subsection{Implementation Details}
For fair comparisons with other methods, we follow \cite{le2020hierarchical} to divide the videos into 8 clips for the SUTD-TrafficQA and TGIF-QA datasets, and $24$ clips for the MSVD-QA and MSRVTT-QA datasets that contain long videos. The Swin-L \cite{liu2021swin} pretrained on ImageNet-22K dataset is used to extract the frame-level appearance features, and the video Swin-B \cite{liu2021videoswin} pretrained on Kinetics-600 is applied to extract the clip-level motion features. For the question, we adopt the pre-trained $300$-dimensional GloVe \cite{pennington2014glove} word embeddings to initialize the word features in the sentence. For parameter settings, we set the dimension $d$ of hidden layer to 512. For the  Multi-modal Transformer Block (MTB), the number of layers $r$ is set to 3 for SUTD-TrafficQA, $8$ for TGIF-QA,  $5$ for MSVD-QA, and $6$ for MSRVTT-QA. The number of attentional heads $H$ is set to 8. The dictionary is initialized by applying K-means over the whole visual features from the whole training set to get $512$ clusters and is updated during end-to-end training. The number of GCN layers $g$ is set to $1$ in the semantic graph embedding. In the training process, we train the model using the Adam optimizer with an initial learning rate 2e-4, a momentum 0.9, and a weight decay 0. The learning rate reduces by half when the loss stops decreasing after every  $5$ epochs. The batch size is set to 64. The dropout rate is set to 0.15 to prevent overfitting. All experiments are terminated after $50$ epochs. We implement our model by PyTorch with an NVIDIA RTX 3090 GPU. For multi-choice and open-ended tasks, we use the accuracy to evaluate the performance of our model. For the counting task in TGIF-QA dataset, we adopt the Mean Squared Error (MSE) between the predicted answer and the right answer.

\subsection{Comparison With State-of-the-Art Methods}
\subsubsection{Results on SUTD-TrafficQA Dataset}
Since the splits of six reasoning tasks are not provided by the original SUTD-TrafficQA dataset \cite{xu2021sutd}, we divide the SUTD-TrafficQA dataset into six reasoning tasks according to the question types. The overall accuracy and the accuracy of each reasoning types are reported. 

The results in Table \ref{Table4} demonstrate that our CMCIR achieves the best performance for six reasoning tasks including basic understanding, event forecasting, reverse reasoning, counterfactual inference, introspection and attribution analysis. Specifically, the CMCIR improves the best state-of-the-art method Eclipse \cite{xu2021sutd} by $1.53\%$ for all reasoning tasks. Compared with the re-implemented methods $\textrm{VQAC}^\dag$, $\textrm{MASN}^\dag$, $\textrm{DualVGR}^\dag$, and $\textrm{HCRN}^\dag$, our CMCIR performs better than these methods in all six tasks by a significant margin. For example, compared with $\textrm{HCRN}^\dag$, our CMCIR improves the accuracy by $1.93\%$ for basic understanding, $2.30\%$ for attribution analysis, $4.98\%$ for introspection, $5.30\%$ for counterfactual inference, $4.22\%$ for event forecasting, $2.12\%$ for reverse reasoning, and $2.32\%$ for all tasks. It is obvious that our method improves three types of questions the most: introspection, counterfactual inference, and event forecasting. The introspection task is to test if models can provide preventive advice that could have been taken to prevent traffic accidents. The event forecasting task is to infer future events based on observed videos, and the forecasting questions inquire about the outcome of the current situation. The counterfactual inference task queries the consequent outcomes of certain hypotheses that did not occur. All of these three question types require causal relational reasoning among the causal, logic, and spatial-temporal structures of the visual and linguistic content. This validates that our CMCIR  can model multi-level interaction and causal relations between the language and spatial-temporal structure of the event-level urban data.

\subsubsection{Results on Other Benchmark Datasets}
To evaluate the generalization ability of  CMCIR on other event-level datasets, we conduct experiments on TGIF-QA, MSVD-QA, and MSRVTT-QA datasets and compare our model with the state-of-the-art methods. The comparison results on \red{the} TGIF-QA dataset are presented in Table \ref{Table5}. We can see that our CMCIR achieves the best performance for \red{the} \emph{Action} and \emph{FrameQA} tasks. Additionally, our CMCIR also achieves relatively high performance for \red{the} \emph{Transition} and \emph{Count} tasks. Specifically, the CMCIR improves the best performing method HAIR \cite{liu2021hair} by $0.3\%$ for the \emph{Action} task, $2.1\%$ for \red{the} \emph{FrameQA} task. For the \emph{Transition} task, the CMCIR also outperforms other comparison methods except CASSG \cite{liu2022cross} and Bridge2Answer \cite{park2021bridge}. For the \emph{Count} task, our CMCIR also achieves a competitive MSE loss value.

\begin{table}[t]\renewcommand\tabcolsep{4.0pt}\renewcommand\arraystretch{1}
\begin{center}
\begin{tabular}{lcccc}
\hline
\multirow{2}*{Method}&\multicolumn{4}{c}{Task Type}\\\cline{2-5}
&Action$\uparrow$&Transition$\uparrow$&FrameQA$\uparrow$&Count$\downarrow$\\\hline
ST-VQA  \cite{jang2017tgif}&62.9&69.4&49.5&4.32\\
Co-Mem  \cite{gao2018motion}&68.2&74.3&51.5&4.10\\
PSAC  \cite{li2019beyond}&70.4&76.9&55.7&4.27\\
HME \cite{fan2019heterogeneous}&73.9&77.8&53.8&4.02\\
GMIN \cite{gu2021graph}&73.0&81.7&57.5&4.16\\
L-GCN \cite{huang2020location}&74.3&81.1&56.3&3.95\\
HCRN \cite{le2020hierarchical}&75.0&81.4&55.9&3.82\\
HGA \cite{jiang2020reasoning}&75.4&81.0&55.1&4.09\\
QueST \cite{jiang2020divide}&75.9&81.0&59.7&4.19\\
Bridge2Answer \cite{park2021bridge}&75.9&\underline{82.6}&57.5&\textbf{3.71}\\
QESAL\cite{liu2021question}&76.1&82.0&57.8&3.95\\
ASTG \cite{jin2021adaptive}&76.3&82.1&\underline{61.2}&\underline{3.78}\\
CASSG \cite{liu2022cross}&77.6&\textbf{83.7}&58.7&3.83\\
HAIR \cite{liu2021hair}&77.8&82.3&60.2&3.88\\\hline
\textbf{CMCIR (ours) }&\textbf{78.1}&82.4&\textbf{62.3}&3.83\\\hline
\end{tabular}
\end{center}
\caption{Comparison with state-of-the-art methods on TGIF-QA dataset. }
\vspace{-20pt}
\label{Table5}
\end{table}

\begin{table}[t]\renewcommand\tabcolsep{3pt}\renewcommand\arraystretch{1}
\begin{center}
\begin{tabular}{lcccccc}
\hline
\multirow{3}*{Method}&\multicolumn{6}{c}{Question Type}\\\cline{2-7}
&What &Who&How&When&Where&All\\
&(8,149)&(4,552)&(370)&(58)&(28)&(13,157)\\\hline
Co-Mem \cite{gao2018motion}&19.6&48.7&81.6&74.1&31.7&31.7\\
AMU \cite{xu2017video}&20.6&47.5&83.5&72.4&\textbf{53.6}&32.0\\
HME \cite{fan2019heterogeneous}&22.4&50.1&73.0&70.7&42.9&33.7\\
HRA \cite{chowdhury2018hierarchical}&-&-&-&-&-&34.4\\
HGA  \cite{jiang2020reasoning}&23.5&50.4&83.0&72.4&46.4&34.7\\
GMIN \cite{gu2021graph}&24.8&49.9&84.1&\underline{75.9}&\textbf{53.6}&35.4\\
QueST \cite{jiang2020divide}&24.5&52.9&79.1&72.4&\underline{50.0}&36.1\\
HCRN \cite{le2020hierarchical}&-&-&-&-&-&36.1\\
CASSG \cite{liu2022cross}&24.9&52.7&\textbf{84.4}&74.1&\textbf{53.6}&36.5\\
QESAL\cite{liu2021question}&25.8&51.7&83.0&72.4&\underline{50.0}&36.6\\
Bridge2Answer \cite{park2021bridge}&-&-&-&-&-&37.2\\
HAIR \cite{liu2021hair}&-&-&-&-&-&37.5\\
VQAC \cite{kim2021video}&26.9&53.6&-&-&-&37.8\\
MASN \cite{seo2021attend}&-&-&-&-&-&38.0\\
HRNAT \cite{gao2022hierarchical}&-&-&-&-&-&38.2\\
ASTG \cite{jin2021adaptive}&26.3&\underline{55.3}&82.4&72.4&\underline{50.0}&38.2\\
DualVGR \cite{wang2021dualvgr}&\underline{28.6}&53.8&80.0&70.6&46.4&\underline{39.0}\\\hline
\textbf{CMCIR (ours)}&\textbf{33.1}&\textbf{58.9}&\underline{84.3}&\textbf{77.5}&42.8&\textbf{43.7}\\\hline
\end{tabular}
\end{center}
\caption{Comparison with state-of-the-art methods on MSVD-QA dataset.  }
\vspace{-20pt}
\label{Table6}
\end{table}

Table \ref{Table6} shows the comparison results on the MSVD-QA dataset. From the results, we can see that our CMCIR outperforms nearly all the state-of-the-art \red{comparison} methods by a significant margin. For example, our CMCIR achieves the best overall accuracy of $43.7\%$, which leads to $4.7\%$ improvement over the best performing method DualVGR \cite{wang2021dualvgr}. For \emph{What}, \emph{Who}, and \emph{When} types, the CMCIR \red{significantly} outperforms all the comparison methods. Although GMIN \cite{gu2021graph} and CASSG \cite{liu2022cross} perform marginally better than our CMCIR for \emph{How} and \emph{Where} types, our CMCIR performs significantly better than GMIN for What ($+8.3\%$), Who ($+9.0\%$), When ($+1.6\%$), and the overall ($+8.3\%$) tasks.

Table \ref{Table7} shows the comparison results \red{for} the MSRVTT-QA dataset. It can be observed that our CMCIR \red{outperforms} the  best performing method ASTG \cite{jin2021adaptive}, with the highest accuracy of $38.9\%$. For \emph{What}, \emph{Who}, and \emph{When} question types,  the CMCIR performs the best compared \red{to} all the previous state-of-the-art methods. Although CASSG \cite{liu2022cross} and GMIN \cite{gu2021graph} achieve better accuracies than our CMCIR for \emph{How} and \emph{Where} question types respectively,  our CMCIR achieves \red{a} significantly performance improvement \red{over} these two methods for other question types.

\begin{table}[!t]\renewcommand\tabcolsep{2.0pt}\renewcommand\arraystretch{1}
\begin{center}
\begin{tabular}{lcccccc}
\hline
\multirow{3}*{Method}&\multicolumn{6}{c}{Question Type}\\\cline{2-7}
&What &Who&How&When&Where&All\\
&(49,869)&(20,385)&(1,640)&(677)&(250)&(72,821)\\\hline
Co-Mem \cite{gao2018motion}&23.9&42.5&74.1&69.0&42.9&31.9\\
AMU \cite{xu2017video}&26.2&43.0&80.2&72.5&30.0&32.5\\
HME \cite{fan2019heterogeneous}&26.5&43.6&82.4&76.0&28.6&33.0\\
QueST \cite{jiang2020divide}&27.9&45.6&83.0&75.7&31.6&34.6\\
HRA \cite{chowdhury2018hierarchical}&-&-&-&-&-&35.0\\
MASN \cite{seo2021attend}&-&-&-&-&-&35.2\\
HRNAT \cite{gao2022hierarchical}&-&-&-&-&-&35.3\\
HGA \cite{jiang2020reasoning}&29.2&45.7&83.5&75.2&34.0&35.5\\
DualVGR \cite{wang2021dualvgr}&29.4&45.5&79.7&76.6&36.4&35.5\\
HCRN \cite{le2020hierarchical}&-&-&-&-&-&35.6\\
VQAC \cite{kim2021video}&29.1&46.5&-&-&-&35.7\\
CASSG \cite{liu2022cross}&29.8&46.3&\textbf{84.9}&75.2&35.6&36.1\\
GMIN \cite{gu2021graph}&30.2&45.4&\underline{84.1}&74.9&\textbf{43.2}&36.1\\
QESAL\cite{liu2021question}&30.7&46.0&82.4&76.1&\underline{41.6}&36.7\\
Bridge2Answer \cite{park2021bridge}&-&-&-&-&-&36.9\\
HAIR \cite{liu2021hair}&-&-&-&-&-&36.9\\
ClipBERT \cite{lei2021less}&-&-&-&-&-&37.4\\
ASTG \cite{jin2021adaptive}&\underline{31.1}&\underline{48.5}&83.1&\underline{77.7}&38.0&37.6\\\hline
\textbf{CMCIR (ours)}&\textbf{32.2}&\textbf{50.2}&82.3&\textbf{78.4}&38.0&\textbf{38.9}\\\hline
\end{tabular}
\end{center}
\caption{Comparison with state-of-the-art methods on MSRVTT-QA dataset. }
\vspace{-20pt}
\label{Table7}
\end{table}

In Table  \ref{Table6} and Table  \ref{Table7}, our method achieves lower performance than previous best method when the question types are  \emph{How} and \emph{Where}. It can be seen from Table  \ref{Table6} and Table  \ref{Table7} that the number of \emph{How} and \emph{Where} samples are much smaller than that of the other question types. Due to the existence of data bias in these two datasets, the model tends to learn spurious correlation from other question types. This may lead to the performance degradation when testing on these two question types. Nonetheless, we can still obtain promising performance for question type \emph{When}, which also has limited samples. This validates that our CMCIR indeed mitigate the spurious correlations for most of the question types including \emph{What}, \emph{Who}, and \emph{When}.

The experimental results in Table \ref{Table5}-\ref{Table7} show that our CMCIR outperforms state-of-the-art methods on three large-scale benchmark event-level datasets. This validates that our CMCIR generalizes well across different event-level datasets, including urban traffic and real-world scenes. Our CMCIR achieves more promising performance than existing relational reasoning methods like HGA, QueST, GMIN, Bridge2Answer, QESAL, ASTG, PGAT, HAIR and CASSG, which validates that our CMCIR has good potential to model multi-level interaction and causal relations between the language and spatial-temporal structure of videos. The main reason for good generalization across different datasets is that our CMCIR can mitigate both the visual and linguistic biases through front-door and back-door causal intervention modules. Due to the strong multi-modal relational reasoning ability of the CMCIR, we can disentangle the spurious correlations within visual-linguistic modality and achieve robust spatial-temporal relational reasoning.

Comparing the average improvement across different datasets, we notice that CMCIR achieves the best improvement on SUTD-TrafficQA (+1.53\%), MSVD-QA (+4.7\%) while relatively moderate gains on TGIF-QA (+0.3\%$\thicksim$0.9\%) and MSRVTT-QA (+1.3\%). The reason for such discrepancy is that SUTD-TrafficQA and MSVD-QA are relatively small in size, which constrains the reasoning ability of the backbone models by limiting their exposure to training instances. As a comparison, SUTD-TrafficQA is four times smaller than MSRVTT-QA in terms of QA pairs (60K vs 243K), while MSVD-QA is five times smaller than MSRVTT-QA in terms of QA pairs (43K vs 243K). However, such deficiency caters to the focal point of our CMCIR, which develops better in a less generalized situation, thus leading to more preferable growth on MSVD-QA. This validates that our causality-aware visual-linguistic representation has good generalization ability.

\begin{table}[!t]\renewcommand\tabcolsep{0.5pt}\renewcommand\arraystretch{1}
\begin{center}
\begin{tabular}{lccccccc}
\hline
&CMCIR &CMCIR&CMCIR&CMCIR&CMCIR&CMCIR&\\
Datasets&w/o&w/o&w/o&w/o&w/o&w/o&CMCIR\\
&HSRP&LBCI&VFCI&CVLR&SGE&ALFF&\\\hline
SUTD&37.65&37.71&37.68&37.42&37.93&37.84&\textbf{38.58}\\
TGIF (Action)&75.4&75.1&75.5&75.0&75.4&75.2&\textbf{78.1}\\
TGIF (Transition)&81.2&81.3&80.6&80.4&81.0&81.2&\textbf{82.4}\\
TGIF (FrameQA)&62.0&61.9&61.6&61.2&61.3&61.1&\textbf{62.3}\\
TGIF (Count)&4.03&3.89&4.10&4.05&3.91&4.12&\textbf{3.83}\\
MSVD&42.4&42.7&42.2&42.0&42.9&42.5&\textbf{43.7}\\
MSRVTT&38.5&38.3&38.1&38.0&38.2&38.4&\textbf{38.9}\\\hline
\end{tabular}
\end{center}
\caption{Ablation study on SUTD-TrafficQA, TGIF-QA, MSVD-QA, and MSRVTT-QA datasets. }
\vspace{-20pt}
\label{Table8}
\end{table}

\subsection{Ablation Studies}

We further conduct ablation experiments using the following variants of CMCIR to verify the contributions of the components designed in out method.
\begin{itemize}
\item CMCIR w/o HSRP:  we remove the Hierarchical Semantic-Role Parser (HSRP),  which parses the question into verb-centered relation tuples (subject, relation, object). The CMCIR model only \red{uses} the original question as the linguistic representation.
\item CMCIR w/o LBCI:  we remove the Linguistic Back-door Causal Intervention (LBCI) module. The CVLR module only contains visual front-door causal intervention (VFCI) module.
\item CMCIR w/o VFCI:  we remove the Visual Front-door Causal Intervention (VFCI) module. The CVLR module only contains linguistic back-door causal intervention (LBCI) module.
\item CMCIR w/o CVLR: we remove the Causality-aware Visual-Linguistic Reasoning (CVLR) module. The CMCIR model combines the visual and linguistic representations using spatial-temporal transformer (STT) and visual-linguistic feature fusion modules.
\item CMCIR w/o SGE: we remove the Semantic Graph Embedding (SGE) module when conducting visual-linguistic feature fusion. The linguistic features are directly used for adaptive linguistic feature fusion.
\item CMCIR w/o ALFF: we remove the Adaptive Linguistic Feature Fusion (ALFF) module when conducting visual-linguistic feature fusion. The semantic graph embedded linguistic features are directly used to fused with the visual features.
\end{itemize}

Table \ref{Table8} shows the evaluation results of the ablation study on SUTD-TrafficQA, TGIF-QA, MSVD-QA, and MSRVTT-QA datasets. It can be observed that our CMCIR achieves the best performance compared to the six variants across all datasets and tasks. Without HSRP, the performance drops significantly due to the lack of the hierarchical linguistic feature representation. This shows that our proposed hierarchical semantic-role parser indeed increase the representation ability of question semantics. To be noticed, the performance of CMCIR w/o LBCI, CMCIR w/o VFCI, and CMCIR w/o CVLR are all lower than that of the CMCIR. This validates that both the linguistic back-door and visual front-door causal interventions contribute to discover the causal structures and learn the causality-aware visual-linguistic representations, and thus improve the model performance. For CMCIR w/o SGE and CMCIR w/o ALFF, their performance are higher than that of the CMCIR w/o LBCI, CMCIR w/o VFCI, and CMCIR w/o CVLR, but lower than that of our CMCIR, which indicates effectiveness of semantic graph embedding and adaptive linguistic feature fusion that leverages the hierarchical linguistic semantic relations as the guidance to adaptively learn the global semantic-aware visual-linguistic representations. With all the components, our CMCIR performs the best because all these components are beneficial and work collaboratively to achieve robust event-level visual question answering.

\begin{table*}[t]\renewcommand\tabcolsep{8pt}\renewcommand\arraystretch{1}
\begin{center}
\begin{tabular}{lcccccccc}
\hline
&&\multirow{2}*{SUTD-TrafficQA}&TGIF-QA&TGIF-QA&TGIF-QA&TGIF-QA&\multirow{2}*{MSVD-QA}&\multirow{2}*{MSRVTT-QA}\\
&&& (Action) &(Transisition) &(FrameQA)&(Count)&&\\\hline
\multirow{5}*{MMA Heads}&1&37.83&75.8&80.7&61.2&3.92&42.3&38.5\\
&2&38.17&75.7&79.7&60.6&3.96&42.0&38.5\\
&4&37.51&75.8&79.2&61.1&3.93&42.2&38.3\\
&8&\textbf{38.58}&\textbf{78.1}&\textbf{82.4}&\textbf{62.3}&\textbf{3.83}&\textbf{43.2}&\textbf{38.9}\\\hline
\multirow{8}*{MTB Layers}&1&37.81&74.5&79.4&60.3&4.26&42.9&38.7\\
&2&37.98&74.8&80.4&61.0&4.20&42.8&38.2\\
&3&\textbf{38.58}&75.1&80.1&61.0&4.03&43.0&38.4\\
&4&37.84&76.6&80.2&61.6&3.96&42.6&38.7\\
&5&37.63&75.5&80.6&61.0&3.94&\textbf{43.7}&38.7\\
&6&37.73&76.2&80.8&61.4&4.12&43.2&\textbf{38.9}\\
&7&37.73&75.4&80.3&61.2&3.98&43.1&38.3\\
&8&37.58&\textbf{78.1}&\textbf{82.4}&\textbf{62.3}&\textbf{3.83}&42.8&38.6\\\hline
\multirow{3}*{GCN Layers}&1&\textbf{38.58}&\textbf{78.1}&\textbf{82.4}&\textbf{62.3}&\textbf{3.83}&\textbf{43.2}&\textbf{38.9}\\
&2&37.84&74.9&80.3&61.0&4.07&41.8&38.3\\
&3&37.58&74.7&80.3&60.8&4.03&42.1&38.4\\\hline
\multirow{3}*{Dimension}&256&37.60&73.9&79.9&61.0&3.96&42.8&38.8\\
&512&\textbf{38.58}&\textbf{78.1}&\textbf{82.4}&\textbf{62.3}&\textbf{3.83}&\textbf{43.2}&\textbf{38.9}\\
&768&37.74&75.0&80.0&62.2&3.90&42.8&38.0\\\hline
\end{tabular}
\end{center}
\caption{Performance of CMCIR with different values of MMA heads, MTB layers, GCN layers, and hidden state dimension on the SUTD-TrafficQA, TGIF-QA, MSVD-QA, and MSRVTT-QA datasets.  }
\vspace{-15pt}
\label{Table9}
\end{table*}

\subsection{Parameter Sensitivity}
To evaluate how the hyper-parameters of CMCIR \red{affect} the performance, we report the results of different values of the heads $h$ of the Multi-head Multi-modal Attention (MMA) module, the layers $r$ of Multi-modal Transformer Block (MTB), and GCN layers $g$ in the semantic graph embedding. Moreover, the dimension of hidden states $d$ is also analyzed. The results \red{for} the SUTD-TrafficQA, TGIF-QA, MSVD-QA, and MSRVTT-QA datasets are shown in Table \ref{Table9}. We can see that the performance of CMCIR with $8$ MMA heads performs the best across all datasets and tasks compared to CMCIR with fewer MMA heads. This indicates that more heads can facilitate the MMA module \red{to} employ more perspectives to explore the relations between different modalities. For MTB layers, the optimal layer numbers are different for different datasets. The performance of the CMCIR is the best when the number of MTB layers is $3$ on \red{the} SUTD-TrafficQA dataset, $8$ on TGIF-QA dataset, $5$ on \red{the} MSVD-QA dataset, and $6$ on \red{the} MSRVTT-QA dataset. For GCN layers, we can see that more GCN layers will increase the amount of learnable parameters and thus make model converge more difficultly. Since one GCN layer can achieve the best performance, we choose \red{one-layer} GCN. For the dimension of hidden states, we can see that $512$ is the best dimensionality of hidden states of the VLICR model due to its good compromise between feature representation ability and model complexity.

\begin{table}[!t]\renewcommand\tabcolsep{2.0pt}\renewcommand\arraystretch{1}
\begin{center}
\begin{tabular}{lcccc}
\hline
&Method&Appearance&Motion&Accuracy\\\hline
\multirow{3}*{SUTD-QA}&Eclipse \cite{xu2021sutd}&ResNet-101&MobileNetV2&37.05\\
&Ours&Swin-L&Video Swin-B&\textbf{38.58} (+1.54)\\
&Ours&ResNet-101&ResNetXt-101&\underline{38.10} (+1.05)\\\hline
\multirow{3}*{MSVD-QA} &DualVGR \cite{wang2021dualvgr}&ResNet-101&ResNetXt-101&39.0\\
&Ours&Swin-L&Video Swin-B&\textbf{43.7} (+4.70)\\
&Ours&ResNet-101&ResNetXt-101&\underline{40.3} (+1.30)\\\hline
\multirow{3}*{MSRVTT-QA} &HCRN \cite{le2020hierarchical}&ResNet-101&ResNeXt-101&35.6\\
&Ours&Swin-L&Video Swin-B&\textbf{38.9} (+3.30)\\
&Ours&ResNet-101&ResNeXt-101&\underline{37.0} (+1.40)\\\hline
\end{tabular}
\end{center}
\caption{Performance of CMCIR with different visual appearance and motion features on SUTD-TrafficQA, MSVD-QA, and MSRVTT-QA datasets. }
\vspace{-25pt}
\label{Table10}
\end{table}

To validate whether our CMCIR \red{can} generalize to different visual appearance and motion features, we evaluate the performance of the CMCIR on \red{the} SUTD-TrafficQA, MSVD-QA and MSRVTT-QA datasets using different visual appearance and motion features, as shown in Table \ref{Table10}. The best performing comparison methods on \red{the} SUTD-TrafficQA, MSVD-QA and MSRVTT-QA datasets are also shown in Table \ref{Table10}. It can be observed that when using Swin-L and Video Swin-B as the visual and motion features, our CMCIR can achieves the state-of-the-art performance compared with other methods. In our experiments, visual appearance features are the pool5 output of ResNet-101\cite{he2016deep} and visual motion features  are derived by ResNetXt-101 \cite{xie2017aggregated,hara2018can}. When using ResNet-101 and ResNetXt-101 as the visual and motion features, our CMCIR can also achieve competitive accuracy on SUTD-TrafficQA, MSVD-QA and MSRVTT-QA datasets. For SUTD-TrafficQA dataset, the performance of using ResNet and ResNetXt is $38.10\%$, which is the also the best accuracy among all the comparison methods (Table \ref{Table4}). For \red{the} MSVD-QA dataset, the performance of using ResNet-101 and ResNetXt-101 is $40.3\%$, which also outperforms other comparison methods (Table \ref{Table6}). For \red{the} MSRVTT-QA dataset, the performance of using ResNet-101 and ResNetXt-101 is $37.0\%$, which also achieves competitive performance \red{compared to} other comparison methods (Table \ref{Table6}). These results validates that our CMCIR generalizes well across different visual appearance and motion features due to the learned causality-aware visual-linguistic representations. More importantly, the performance improvement of our CMCIR is mainly attributed to our elaborately designed visual-linguistic causal reasoning model.

\subsection{The Evidence of Reducing Spurious Correlations}
\red{Actually, the process of building VideoQA datasets will introduce undesirable spurious correlations rather than the overarching reality \cite{nan2021interventional}. Therefore, we can assume that all our evaluation datasets contain spurious correlations. To validate the effectiveness of the CVLR module in reducing spurious correlations in non-causal frameworks, we apply the CVLR to three state-of-the-art models Co-Mem \cite{gao2018motion}, HGA \cite{jiang2020reasoning} and HCRN \cite{le2020hierarchical}. Since our CVLR is orthogonal to the backbone, we can insert the CVLR directly after the feature extraction layers of these models, which is the same as our CMCIR. As shown in Table \ref{Table11},  our  CVLR brings each backbone a sharp gain across all benchmark datasets (+0.9\%$\thicksim$6.5\%), which evidences its model-agnostic property. Nevertheless, we notice that the improvements fluctuate across the backbones. As a comparison, on MSVD-QA and MSRVTT-QA benchmarks, CVLR acquires more favorable gains with backbones Co-Mem, HGA and HCRN than it does with our backbone. This is because the fine-grained interactions between linguistic semantics and spatial-temporal representations empower our backbone with robustness, especially to questions of the descriptive type on MSVD-QA and MSRVTT-QA benchmarks. Therefore, it achieves stronger backbone performances on benchmarks that focus on the descriptive question (i.e., MSVD-QA and MSRVTT-QA), which, in turn, accounts for the contribution of CVLR to some extent, thus makes improvement of our backbone less remarkable. In contrast, when it comes to the causal and temporal question (i.e., SUTD-TrafficQA), the CVLR shows equivalent improvements on all four backbones (+1.05\%$\thicksim$2.02\%). These results validate that our CVLR is effective in capturing the causality and reducing the spurious correlations across different models.}

\begin{table}[!t]\renewcommand\tabcolsep{2.0pt}\renewcommand\arraystretch{1}
\begin{center}
\begin{tabular}{lccc}
\hline
Models&SUTD-TrafficQA&MSVD-QA&MSRVTT-QA\\\hline
Co-Mem \cite{gao2018motion}&35.10&34.6&35.3\\
Co-Mem \cite{gao2018motion}+ CVLR&\textbf{37.12} (+2.02)&\textbf{40.7} (+6.1)&\textbf{38.0} (+2.7)\\\hline
HGA \cite{jiang2020reasoning}&35.81&35.4&36.1\\
HGA \cite{jiang2020reasoning}+ CVLR&\textbf{37.23} (+1.42)&\textbf{41.9} (+6.5)&\textbf{38.2} (+2.1)\\\hline
HCRN \cite{le2020hierarchical}&36.49&36.1&35.6\\
HCRN \cite{le2020hierarchical}+ CVLR&\textbf{37.54} (+1.05)&\textbf{42.2} (+6.1)&\textbf{37.8} (+2.2)\\\hline
Our Backbone&37.42&42.0&38.0\\
Our Backbone + CVLR&\textbf{38.58} (+1.16)&\textbf{43.7} (+1.7)&\textbf{38.9} (+0.9)\\\hline
\end{tabular}
\end{center}
\caption{\red{The CVLR module is applied to different existing non-causal models.}}
\vspace{-25pt}
\label{Table11}
\end{table}

\begin{figure*}[t]
\begin{center}
\includegraphics[scale=0.24]{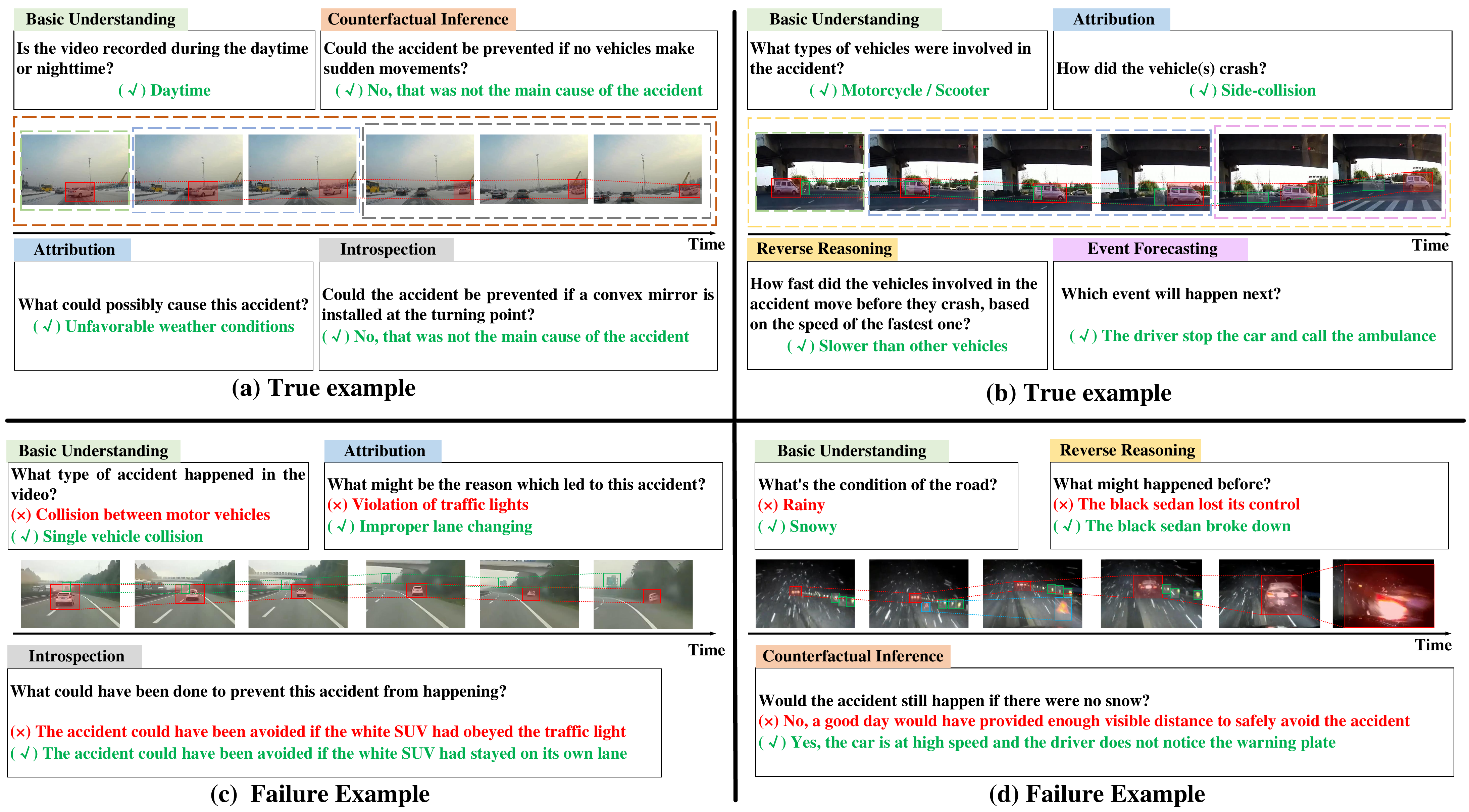}
\end{center}
   \vspace{-15pt}
   \caption{\red{Visualization of visual-linguistic causal reasoning examples from the SUTD-TrafficQA dataset. Each video is accompanied by several question types that contain spurious correlations. The color windows in the videos denote the concentrated visual concepts for the inference.}}
      \vspace{-10pt}
\label{Fig9}
\end{figure*}

\subsection{Qualitative Results }
\red{To verify the ability of the CMCIR in robust spatial-temporal relational reasoning, we aim to gain insight into its visual-linguistic causal reasoning capabilities by inspecting some correct and failure examples from the SUTD-TrafficQA dataset and show the visualization results in Fig. \ref{Fig9}. We show how our model conducts robust spatial-temporal relational reasoning and reduces spurious correlations.}

\red{\textbf{Reliable reasoning}. As shown in Fig. \ref{Fig9} (a), there exists an ambiguity problem where the dominant visual regions of the accident may be distracted by other visual concepts (i.e. different cars/vehicles on the road). In our CMCIR, we learn the question-relevant visual-linguistic association by causal relational learning, thus mitigating such ambiguity in our inference results where video-question-answer triplets exhibit a strong correlation between the dominant spatial-temporal scenes and the question semantics. This validates that the CMCIR can reliably focus on the correct visual regions when making decisions.}

\red{\textbf{Reducing spurious correlation}. In Fig. \ref{Fig9} (b), we present a case reflecting the spurious correlation, where the visual regions of ``van" are spuriously correlated with associated with the ``sedan", due to their frequent co-occurrences. In other words, the model without explicitly considering reducing spurious correlations (e.g., Co-Mem \cite{gao2018motion}, HGA \cite{jiang2020reasoning} and HCRN \cite{le2020hierarchical}) will hesitate when encountering the visual concepts of ``van" and ``motorbike" with regard to region-object correspondence. In our CMCIR, we reduce such spurious correlation and pursue the true causality by adopting visual-linguistic causal intervention, resulting in better dominant visual evidence and question intention. }

\red{\textbf{Generalization ability}. From Fig. \ref{Fig9} (a)-(b), we can see that the CMCIR can generalize well across different question types. which shows that the CMCIR is sensitive to questions and can effectively capture the dominant spatial-temporal content in the videos by conducting robust and reliable spatial-temporal relational reasoning.}

\red{\textbf{Introspective and counterfactual learning}. For challenging question types, such as introspection and counterfactual inference, the CMCIR model can accurately determine whether the attended scene reflects the logic behind the answer. This verifies that the CMCIR can fully explore the causal, logical, and spatial-temporal structures of the visual and linguistic content, due to its promising ability to perform robust visual-linguistic causal reasoning that disentangles visual-linguistic spurious correlations.}

\red{\textbf{Additional failure cases. } Moreover, we provide failure examples in Fig. \ref{Fig9} (c)-(d) to gain further insights into the limitations of our method. In Fig. \ref{Fig9} (c), our model mistakenly correlates the visual concept ``suv" with the green ``traffic plate" when conducting visual-linguistic reasoning. This is because the visual region of ``traffic plate" looks like the ``truck", while only the white ``suv" exists in the video. In Fig. \ref{Fig9} (d), it is difficult to distinguish between ``rainy" and ``snowy" due to their similar visual appearance in the video. Additionally, the ``reflective stripes" along the road are mistakenly considered as the dominant visual concepts. As our CMCIR model lacks an explicit object detection pipeline, some visually ambiguous concepts are challenging to determine. Moreover, without external prior knowledge of traffic rules, some questions such as ``how to prevent the accident" and ``the cause of the accident" are difficult to answer. One possible solution may be to incorporate object detection and external knowledge of traffic rules into our method, which we will explore in our future work.}

\section{Conclusion}
We propose an event-level visual question answering framework named Cross-Modal Causal RelatIonal Reasoning (CMCIR), to mitigate the spurious correlations and discover the causal structures for visual-linguistic modality. To uncover causal structures for visual and linguistic modalities, we propose a Causality-aware Visual-Linguistic Reasoning (CVLR) module, which leverages front-door and back-door causal interventions to disentangle the spurious correlations between visual and linguistic modalities. Extensive experiments on the event-level urban dataset SUTD-TrafficQA and three benchmark real-world datasets TGIF-QA, MSVD-QA, and MSRVTT-QA demonstrate the effectiveness of CMCIR in discovering visual-linguistic causal structures and achieving robust event-level visual question answering. Unlike previous methods that simply eliminate either the linguistic or visual bias without considering cross-modal causality discovery, we apply front-door and back-door causal intervention modules to discover cross-modal causal structures.

\red{We believe our work could shed light on exploring new boundaries of causal analysis in vision-language tasks (\textbf{Causal-VLReasoning}\footnote{Our visual-linguistic causal learning framework https://github.com/HCPLab-SYSU/Causal-VLReasoning, which is a python open-source framework that implements state-of-the-art causal discovery algorithms for visual-linguistic reasoning tasks, such as VQA, Image/Video Captioning, Medical Report Generation, etc.}). In the future, we will further explore more comprehensive causal discovery methods to discover the question-critical scene elements in event-level visual question answering, particularly in the temporal aspect. By further exploiting the fine-grained temporal consistency in videos, we may achieve a model that pursues better causality. Additionally, we can leverage object-level causal relational inference to alleviate the spurious correlations from object-centric entities. Besides, we will incorporate external expert knowledge into our intervention process. Moreover, due to the inherent unobservable nature of properties, how to quantitatively analyze spurious correlations within datasets remains a challenging problem. Thus, we will discover more intuitive and reasonable metrics to compare the effectiveness of different methods in reducing spurious correlations.}

\ifCLASSOPTIONcompsoc
  \section*{Acknowledgments}
\else
  \section*{Acknowledgment}
\fi

This work is supported in part by the National Key R\&D Program of China under Grant No.2021ZD0111601, in part by the National Natural Science Foundation of China under Grant No.62002395 and No.61976250, in part by the Guangdong Basic and Applied Basic Research Foundation under Grant No.2023A1515011530, No.2021A1515012311 and No.2020B1515020048, and in part by the Guangzhou Science and Technology Planning Project under Grant No. 2023A04J2030.

\ifCLASSOPTIONcaptionsoff
  \newpage
\fi

\bibliographystyle{IEEEtran}
\bibliography{IEEEabrv,bibfile}
\vspace{-20pt}
\begin{IEEEbiography}[{\includegraphics[width=1in,height=1.25in,clip,keepaspectratio]{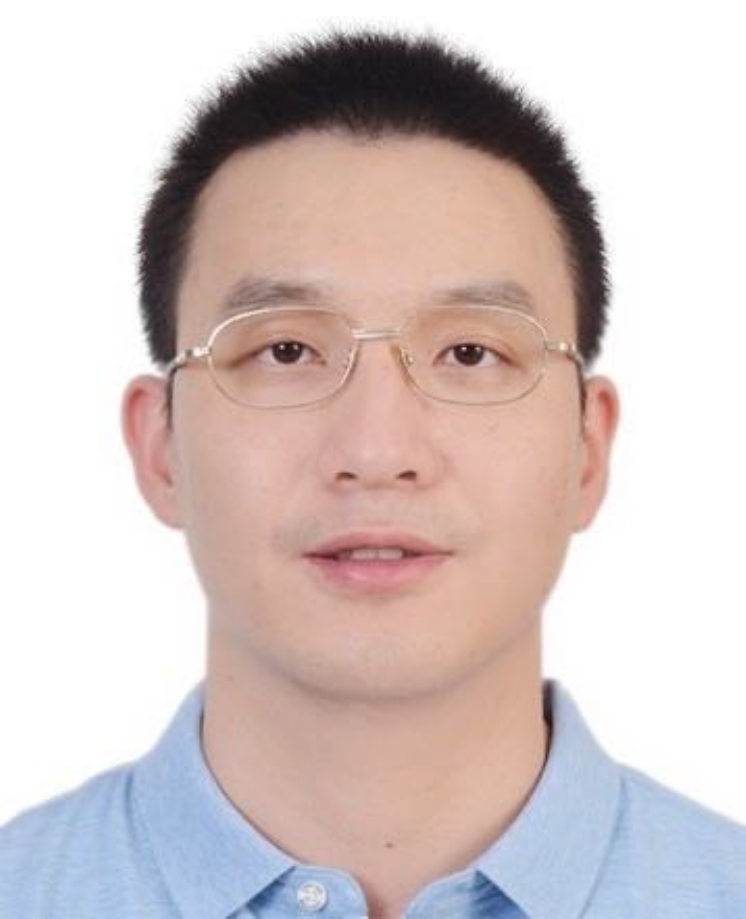}}]{Yang Liu}(M'21) is currently a research associate professor working at the School of Computer Science and Engineering, Sun Yat-sen University. He received his Ph.D. degree from Xidian University in 2019. His current research interests include multi-modal cognitive reasoning and causal relation discovery. He is the recipient of the First Prize of the Third Guangdong Province Young Computer Science Academic Show. He has authorized and co-authorized more than 20 papers in top-tier academic journals and conferences. He has been serving as a reviewer for numerous academic journals and conferences such as TPAMI, TIP, TNNLS, TMM, TCSVT, CVPR, ICCV, ECCV, AAAI, and ACM MM. More information can be found on his personal website https://yangliu9208.github.io.
\end{IEEEbiography}
\vspace{-20pt}
\begin{IEEEbiography}[{\includegraphics[width=1in,height=1.25in,clip,keepaspectratio]{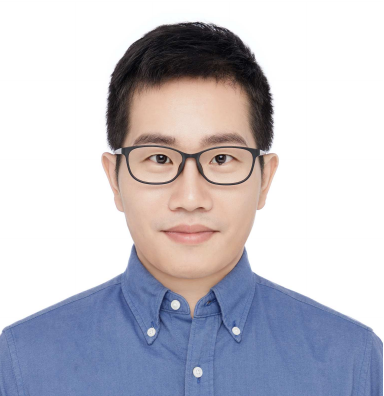}}]{Guanbin Li}(M'15) is currently an associate professor in School of Computer Science and Engineering, Sun Yat-Sen University. He received his PhD degree from the University of Hong Kong in 2016. His current research interests include computer vision, image processing, and deep learning. He is a recipient of ICCV 2019 Best Paper Nomination Award. He has authorized and co-authorized on more than 100 papers in top-tier academic journals and conferences. He serves as an area chair for the conference of VISAPP. He has been serving as a reviewer for numerous academic journals and conferences such as TPAMI, IJCV, TIP, TMM, TCyb, CVPR, ICCV, ECCV and NeurIPS.
\end{IEEEbiography}
\vspace{-20pt}
\begin{IEEEbiography}[{\includegraphics[width=1in,height=1.25in,clip,keepaspectratio]{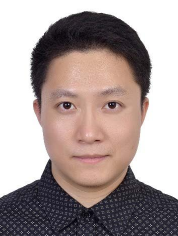}}]{Liang Lin}(M'09, SM'15) is a Full Professor of computer science at Sun Yat-sen University. He served as the Executive Director and Distinguished Scientist of SenseTime Group from 2016 to 2018, leading the R\&D teams for cutting-edge technology transferring. He has authored or co-authored more than 200 papers in leading academic journals and conferences, and his papers have been cited by more than 26,000 times. He is an associate editor of IEEE Trans.Neural Networks and Learning Systems and IEEE Trans. Multimedia, and served as Area Chairs for numerous conferences such as CVPR, ICCV, SIGKDD and AAAI. He is the recipient of numerous awards and honors including Wu Wen-Jun Artificial Intelligence Award, the First Prize of China Society of Image and Graphics, ICCV Best Paper Nomination in 2019, Annual Best Paper Award by Pattern Recognition (Elsevier) in 2018, Best Paper Dimond Award in IEEE ICME 2017, Google Faculty Award in 2012. His supervised PhD students received ACM China Doctoral Dissertation Award, CCF Best Doctoral Dissertation and CAAI Best Doctoral Dissertation. He is a Fellow of IET/IAPR.
\end{IEEEbiography}
\end{document}